\pdfoutput=1
\documentclass[10pt,twocolumn,letterpaper]{article}

\usepackage[T1]{fontenc}
\PassOptionsToPackage{dvipsnames,table}{xcolor}
\usepackage[pagenumbers]{cvpr}      

\usepackage{currfile}
\usepackage{caption}
\usepackage{booktabs}
\usepackage{multirow}
\usepackage{makecell}
\usepackage{amsmath,amssymb,amsthm}
\usepackage{mathtools}
\usepackage{bm}
\usepackage{xspace}
\usepackage{pifont}
\usepackage{colortbl}\usepackage{xcolor}
\usepackage{enumitem}
\setlist{nosep,leftmargin=*}
\setlist[itemize,enumerate]{itemsep=1.2pt,topsep=2.0pt,parsep=0pt,partopsep=0pt,leftmargin=*}

\AtBeginDocument{%
  \setlength{\abovedisplayskip}{5.4pt plus 1.5pt minus 1.0pt}%
  \setlength{\belowdisplayskip}{5.4pt plus 1.5pt minus 1.0pt}%
  \setlength{\abovedisplayshortskip}{2.0pt plus 1.0pt}%
  \setlength{\belowdisplayshortskip}{3.0pt plus 1.0pt}%
}

\newtheorem{lemma}{Lemma}
\newtheorem{proposition}{Proposition}

\theoremstyle{remark}

\newcommand{\ours}{JiL\xspace}
\newcommand{\oursfull}{Judgment-in-the-Loop\xspace}
\definecolor{oursblue}{HTML}{2a78d6}
\definecolor{repairorange}{HTML}{eb6834}
\definecolor{privaqua}{HTML}{1baf7a}
\newcommand{\best}[1]{\textbf{#1}}

\newcommand{\cmark}{\ding{51}}

\newcommand{\E}{\mathbb{E}}

\newcommand{\Q}{Q}
\newcommand{\1}{\mathbf{1}}

\definecolor{cvprblue}{rgb}{0.21,0.49,0.74}
\usepackage[pagebackref,breaklinks,colorlinks,allcolors=cvprblue]{hyperref}

\title{Calibrating World Action Models:\\From Adaptive System~2 Takeover to Self-Evolution}

\author{
Wenhao Li$^{1,4}$ \quad Yu Han$^{2,4}$ \quad Xiu Su$^{3}$ \quad Dehao Wu$^{3}$\\[2pt]
Keke Huang$^{3}$ \quad Shan You$^{4}$ \quad Fei Wang$^{4}$ \quad Chang Xu$^{1}$\\[5pt]
{\normalsize $^{1}$University of Sydney \qquad $^{2}$University of California, San Diego}\\[-1pt]
{\normalsize $^{3}$Central South University \qquad $^{4}$ACE Robotics}
}

\begin{document}
\crefname{assumption}{Assumption}{Assumptions}\Crefname{assumption}{Assumption}{Assumptions}
\crefname{proposition}{Prop.}{Props.}\Crefname{proposition}{Proposition}{Propositions}
\crefname{lemma}{Lemma}{Lemmas}\Crefname{lemma}{Lemma}{Lemmas}\crefname{definition}{Def.}{Defs.}\Crefname{definition}{Definition}{Definitions}
\crefname{corollary}{Cor.}{Cors.}\Crefname{corollary}{Corollary}{Corollaries}
\twocolumn[{%
\renewcommand\twocolumn[1][]{#1}%
\vspace{-2.0em}
\maketitle
\vspace{-2.5em}
\begin{center}
  \centering
  \includegraphics[width=\textwidth]{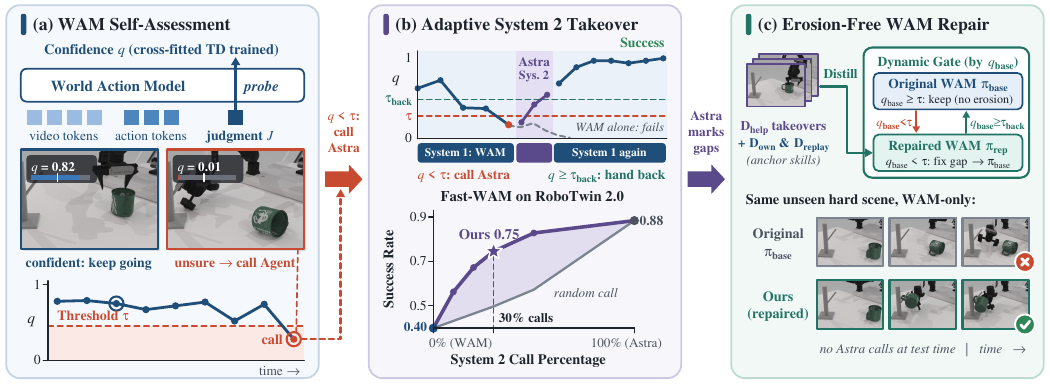}
  \vspace{-1.35em}
  \captionof{figure}{\textbf{\oursfull (\ours).}
  \textbf{(a)}~A single non-invasive judgment token equips a frozen WAM with a calibrated estimate $q$ of its own success probability without affecting its action predictions.
  \textbf{(b)}~Each episode starts with the fast WAM (System~1); when $q$ drops below $\tau$, control switches to a GPT-6-Astra-based agent (System~2) to resolve the bottleneck from the current state, and returns to the WAM once confidence recovers.
  \textbf{(c)}~Because the agent intervenes only when the WAM lacks confidence, its takeover trajectories pinpoint the WAM's capability gaps; distilling these trajectories into the WAM repairs those gaps so it can solve hard scenes without calling Astra at test time, while mastered-skill anchor data ($\mathcal{D}_{\mathrm{own}}, \mathcal{D}_{\mathrm{replay}}$) and $q_{\mathrm{base}}$-based dynamic gating prevent degrading existing capabilities.}
  \label{fig:teaser}
\end{center}
\vspace{-0.6em}
}]

\begin{abstract}
World action models (WAMs) deliver fast visuomotor control, yet degrade sharply in challenging scenes and act blindly when about to fail; conversely, foundation-model-based agents generalize and recover reliably, but incur prohibitive latency and cost when queried at every step.
To achieve both WAM (System 1)-level speed and agent (System 2)-level reliability--- we propose \textbf{\oursfull (\ours)}, a framework that calibrates WAMs' confidence--- adaptively triggering System~2 only when necessary, and distilling System~2's knowledge to form an evolving System~1.
First, \ours equips a frozen WAM with \textbf{calibrated self-assessment}: a single lightweight judgment token reading internal video--action features is trained via our proposed \textbf{cross-fitted temporal-difference (TD) learning} using only binary rollout success/failure signals. It can predict a \textbf{calibrated success probability} at negligible overhead without affecting action predictions.
Second, this calibrated success probability schedules control chunk by chunk---invoking System~2 to resolve bottlenecks when confidence drops, and handing control back to the WAM once confidence recovers, preserving WAM-like speed while approaching System~2 reliability.
Third, as System~2 intervenes only when the WAM lacks confidence, its takeover trajectories directly capture the WAM's capability gaps. JiL distills these takeovers into the WAM to repair those gaps so System~1 can operate standalone without System~2 at test time, while our mastered-skill anchoring and dynamic confidence gating prevent degrading existing capabilities.
Across Fast-WAM, LingBot-VA, and Kairos on RoboTwin~2.0, LIBERO-Plus, and real-world manipulation, \ours System~1 \& 2 outperforms all baselines---e.g., cutting Fast-WAM failures on RoboTwin~2.0 by $57.6\%$ at a $30\%$ agent budget (vs.\ $16.3\%$ for random calls)---while our consolidated standalone WAM improves success rates by $+13.1\%$ to $+24.7\%$ with zero agent calls.

\end{abstract}

\vspace{-10pt}
\section{Introduction}
\label{sec:intro}
Generalist robotic manipulation has recently advanced rapidly along two complementary paradigms. On the reactive side, World action models (WAMs)~\cite{unipi,uva,cosmospolicy,lingbotva,fastwam,wang2026openwam,kairos} couple action generation with world modeling for fast visuomotor control~\cite{robotwin2,libero}, yet degrade sharply in challenging OOD scenes and lack self-assessment---blindly executing erroneous actions even in scenarios beyond their capability~\cite{robotwinplus,liberoplus}. On the deliberative side, manipulation agents built on frontier foundation models such as GPT-6-Astra~\cite{jia2026agp,zhang2026astra,he2026astra,liu2026roboicl,galbot2026astra} offer high task capability, broad generalization, and reliable error recovery, but incur high latency and token cost when queried at every step. Bridging a fast WAM (System~1) with a foundation-model-based agent (System~2) (\cref{fig:teaser}) promises the best of both worlds, yet faces two fundamental bottlenecks: (1)~the \emph{Scheduling Gap} and (2)~the \emph{Consolidation Gap}.

The \emph{Scheduling Gap} lies in knowing \emph{when} to invoke System~2: calling a slow agent only on impending failures requires a fast, calibrated estimate of System~1's success probability before executing each chunk. Some existing methods achieve this estimation by querying external VLMs~\cite{grislain2026ifailsense,spotter}, but incur heavy latency and miss System~1's internal action intent. Others compute multi-sample consistency~\cite{sentinel,vlacalib,vgps} or simulate world-model rollouts ~\cite{foresight,geotts,fare,favor}, which require repeated forward passes or future generation, sacrificing reactive speed. Single-pass feature and conformal detectors~\cite{knowno,conformal,safe,vlaconf,seligmann2026vlafail} instead output uncalibrated anomaly alarms~\cite{knowno,conformal,safe,seligmann2026vlafail} or action-agnostic observation scores~\cite{vlaconf}, rather than calibrated task-success probabilities. Meanwhile, the \emph{Consolidation Gap} lies in turning System~2's help into permanent System~1 capability: existing dual-system frameworks~\cite{zhang2026harness,spotter,callastra,galbot2026astra} treat System 2 as a permanent runtime dependency, repeatedly paying its high latency and cost without internalizing its experience into System 1.

To bridge both gaps, we propose \textbf{\oursfull (\ours)}, a self-evolving System~1 \& 2 framework that equips the WAM with calibrated self-assessment---enabling adaptive System~2 takeover at runtime and continual skill consolidation into System~1.
To resolve the Scheduling Gap, \ours appends a lightweight \emph{judgment token} to the frozen WAM's final denoising step via a causal mask and token-gated adapters~\cite{lora}, reading internal video--action features without affecting action predictions. Trained solely on binary rollout success/failure signals via our \emph{cross-fitted temporal-difference (TD) learning}, which propagates multi-step value estimates without self-bootstrapping bias, this token outputs a calibrated success probability in only $1.6$\,ms per chunk.
Guided by this probability, \ours runs the WAM by default, invokes System~2 below a confidence threshold, and returns control to the WAM once confidence recovers.

To resolve the Consolidation Gap, \ours progressively distills knowledge from System~2's takeovers into System~1, turning runtime rescues into permanent skills and eliminating the need to call System~2 at test time. Because System~2 intervenes only when the WAM lacks confidence, these takeover trajectories automatically pinpoint System~1's capability gaps without manual labeling. However, we prove via a \emph{competence shrinkage law} (\cref{sec:method:gate}) that directly fine-tuning the base WAM on off-distribution recovery data pulls it away from its original behavior, fixing those gaps at the cost of hurting mastered skills. Guided by this law, \ours keeps the base WAM frozen, trains a separate repaired action expert on System~2 takeovers mixed with mastered-skill anchor data (own successes and pre-training replay), and reuses the base WAM's calibrated confidence to switch to the repaired expert only on low-confidence states---preserving mastered skills while achieving multi-round self-evolution.

Our main contributions are as follows:
\begin{itemize}
\item We propose \textbf{\oursfull (\ours)}, a two-system framework uniting WAM speed with agent reliability---slashing agent calls via on-demand takeover and evolving System~1 to run standalone.
  \item We design a non-invasive \textbf{judgment token} trained via \textbf{cross-fitted TD learning} on binary outcomes alone, predicting unbiased, calibrated success probabilities in $1.6$\,ms per chunk without altering actions.
  \item We prove a \textbf{competence shrinkage law} revealing the repair--erosion trade-off, guiding mastered-skill anchoring and \textbf{dynamic confidence gating} for erosion-free gap repair and self-evolution.
  \item Across Fast-WAM, LingBot-VA, and Kairos on RoboTwin~2.0, LIBERO-Plus, and real-world tasks, \ours beats all $30\%$-call baselines and boosts standalone WAM success by $+13.1\%$ to $+24.7\%$ with zero agent calls.
\end{itemize}

\section{Related Work}
\label{sec:related}

\paragraph{World action models.}
WAMs couple video diffusion backbones~\cite{wan,unipi} with action generation to model physical dynamics and future world states: UVA and Cosmos Policy jointly predict future video and actions~\cite{uva,cosmospolicy}, LingBot-VA unifies autoregressive video dynamics and causal action generation in a mixture-of-transformers~\cite{lingbotva,mot}, OpenWAM systematizes modular world--action pretraining~\cite{wang2026openwam}, and Fast-WAM bypasses test-time video imagination for low-latency control~\cite{fastwam}.
By leveraging world modeling and future prediction, WAMs achieve strong dexterous manipulation performance~\cite{robotwin2,libero}, yet remain fragile in challenging out-of-distribution scenes~\cite{robotwinplus,liberoplus}.

\paragraph{Confidence prediction in embodied models.}
Runtime success prediction for embodied policies follows four main paradigms:
(i)~\emph{external VLM failure judges}~\cite{grislain2026ifailsense,spotter}, which incur high latency and miss internal action intent;
(ii)~\emph{multi-sample consistency, prompt ensembling, and candidate scoring}~\cite{sentinel,vlacalib,vgps}, which require multiple forward passes or discrete action-token logits;
(iii)~\emph{world-model rollout verification}~\cite{foresight,geotts,fare,favor}, which sacrifices reactive speed by generating future latents or videos at test time; and
(iv)~\emph{internal-feature and conformal detectors}~\cite{knowno,conformal,safe,vlaconf,seligmann2026vlafail}, which either output uncalibrated anomaly alarms and abstention sets~\cite{knowno,conformal,safe,seligmann2026vlafail} or post-hoc calibrate action-agnostic observation features~\cite{vlaconf}.
In contrast, our proposed cross-fitted TD judgment token calibrates to the accurate task-success probability at negligible overhead without affecting action predictions, enabling both on-demand System~2 takeover and erosion-free self-evolution.

\paragraph{Foundation-model-based embodied control agents.}
Foundation-model-based manipulation agents divide into two tracks.
\emph{Pure foundation-model-based agents} drive robots directly without learned visuomotor policies, spanning programmatic and constraint-synthesis agents~\cite{rekep,liu2026guava,chen2026gap}, zero-shot direct and tool-mediated controllers powered by GPT-6-Astra~\cite{jia2026agp,zhang2026astra,he2026astra,zhou2026gptgear}, and in-context memory-augmented agents~\cite{cheng2026gptpolicy,guo2026robodawn,liu2026roboicl}, all of which face high per-step latency and token cost~\cite{galbot2026astra,zhou2026gptgear}.
\emph{Dual-system frameworks} pair foundation-model-based agents with embodied policies via memory-guided primitive orchestration~\cite{zhang2026harness}, parallel VLM screening and primitive-based failure repair~\cite{spotter}, learned post-hoc assistance heads or proposal-review gates for selective agent hand-offs~\cite{callastra,galbot2026astra}, or agent-guided policy distillation that replaces the base model wholesale~\cite{liu2026guava}.
However, keeping the agent permanently in the runtime loop perpetuates high invocation costs, while wholesale replacement causes negative flips and degrades already-learned skills~\cite{bct,pct,wiseft,biderman2024lora}.
\ours bridges both tracks by invoking System~2 adaptively only when the WAM lacks confidence and consolidating those takeover trajectories into a standalone WAM without degrading existing capabilities.

\begin{figure*}[t]
  \centering
  \includegraphics[width=\textwidth]{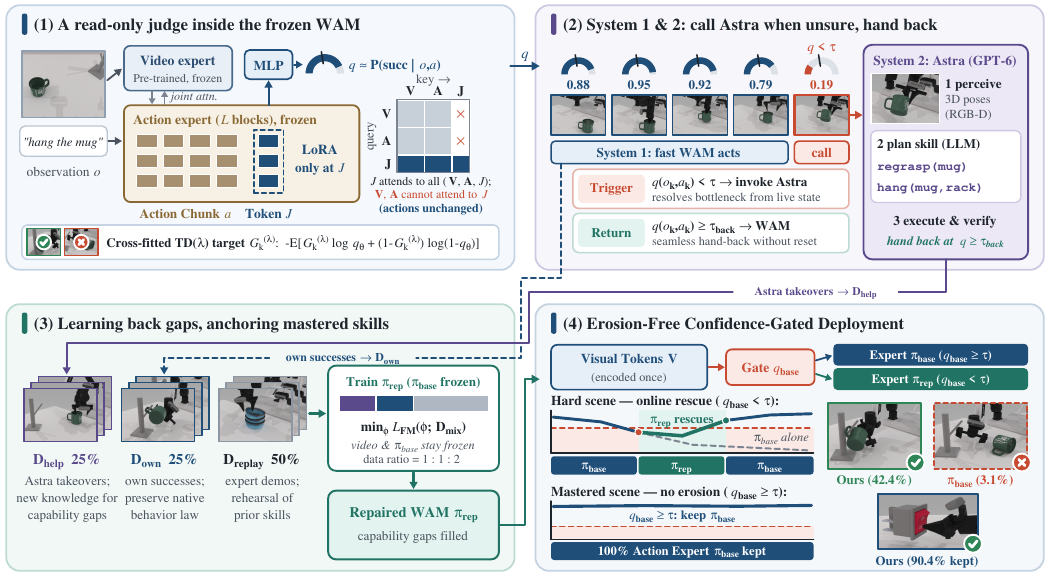}
  \caption{\textbf{Overview of the \oursfull (\ours) framework.}
  (1)~\textbf{Non-Invasive Calibrated Self-Assessment (\cref{sec:method:token}):} A lightweight judgment token $J$ attends to the frozen WAM's internal video and action streams through a causal mask and token-gated adapters, predicting calibrated success probability $q_\theta$ via our proposed cross-fitted TD learning without affecting action predictions.
  (2)~\textbf{Adaptive System~1 \& 2 Takeover (\cref{sec:method:defer}):} At each decision step, $q_\theta$ schedules control chunk by chunk; the WAM hands control to a GPT-6-Astra-based agent when $q_\theta<\tau$ and reclaims execution once confidence recovers above $\tau_{\mathrm{back}}$.
  (3)~\textbf{Capability-Gap Consolidation (\cref{sec:method:distill}):} Keeping $\pi_{\mathrm{base}}$ frozen, successful agent takeovers ($\mathcal{D}_{\mathrm{help}}$) are combined with mastered-skill anchor data---the WAM's own confident successes ($\mathcal{D}_{\mathrm{own}}$) and pre-training replay ($\mathcal{D}_{\mathrm{replay}}$)---into $\mathcal{D}_{\mathrm{mix}}$ to train a separate repaired action expert $\pi_{\mathrm{rep}}$, distilling System~2's recovery skills into System~1.
  (4)~\textbf{Erosion-Free Confidence-Gated Deployment (\cref{sec:method:gate}):} During standalone deployment, $\pi_{\mathrm{base}}$ dynamically switches to $\pi_{\mathrm{rep}}$ when base confidence $q_{\mathrm{base}}$ drops below $\tau$ and reclaims control once confidence recovers, repairing hard states without degrading mastered skills or invoking System~2 (scores shown: Fast-WAM on RoboTwin~2.0, 1 round).}
  \label{fig:method}
\end{figure*}

\section{Method}
\label{sec:method}
\ours couples a frozen WAM (System~1) with a foundation-model-based agent (System~2) through a calibrated internal judgment signal (\cref{fig:method}), linking calibrated self-assessment (\cref{sec:method:token}), adaptive System~2 takeover (\cref{sec:method:defer}), capability-gap consolidation (\cref{sec:method:distill}), and erosion-free confidence-gated deployment and self-evolution (\cref{sec:method:gate,sec:method:loop}).

\subsection{Preliminaries: World Action Models}
\label{sec:method:prelim}
At decision step $k$, a visuomotor policy maps observation $o_k$ (camera images, proprioception, and language instruction) to an action chunk $a_k\in\mathbb{R}^{H\times d}$ over horizon $H$; an episode $\xi=(o_0,a_0,\dots,o_{T-1},a_{T-1})$ of length $T{=}|\xi|$ ends with binary outcome $Y(\xi)\in\{0,1\}$.
Modern WAMs~\cite{mot,fastwam,lingbotva,kairos} use a mixture-of-transformers coupling a pre-trained video expert with an action expert.
At layer $l\in\{1,\dots,L\}$, the video expert encodes $o_k$ into visual tokens $V^{(l)}$, while the action expert updates action tokens $A^{(l)}$ (including proprioceptive tokens) over $Z^{(l)}{=}[V^{(l)};A^{(l)}]$ (with projections $Q_l,K_l,W^V_l,O_l$ and feed-forward weights $W^{\mathrm{up}}_l,W^{\mathrm{down}}_l$):
\begin{equation}
  A^{(l+1)}=\mathrm{FFN}_l\!\big(\mathrm{Attn}(Q_l A^{(l)},\,K_l Z^{(l)},\,W^V_l Z^{(l)})\big).
  \label{eq:mot}
\end{equation}
The action expert parameterizes a flow-matching velocity field $v_\phi$~\cite{flowmatching} transporting noise $\varepsilon\sim\mathcal{N}(0,I)$ to clean actions along $a^{t}=(1-t)\varepsilon+t\,a$, trained on dataset $\mathcal{D}$ via
\begin{equation}
  \mathcal{L}_{\mathrm{FM}}(\phi;\mathcal{D})=\E_{(o,a)\sim\mathcal{D},\,t,\,\varepsilon}\big\|v_\phi(a^{t},t\mid o)-(a-\varepsilon)\big\|^2 .
  \label{eq:fm}
\end{equation}
Despite fast visuomotor control, the generative policy lacks self-assessment to tell whether its sampled chunk $a_k$ will succeed; below, we read \eqref{eq:mot} for calibrated self-assessment and optimize \eqref{eq:fm} to consolidate recovery skills.

\subsection{Non-Invasive Calibrated Self-Assessment}
\label{sec:method:token}
To anticipate impending failures, we learn an online estimator $q_\theta(o_k,a_k)\approx P(Y{=}1\mid o_k,a_k;\pi_{\mathrm{base}})$ predicting the base policy's task-success probability given $(o_k,a_k)$ without affecting its action predictions.

\paragraph{Causal judgment token.}
During the final ($N$-th) Euler denoising step ($t{=}t_{N-1}$, whose velocity prediction deterministically produces $a_k$), we reuse cached visual key--value states and append a single learnable judgment token $j^{(0)}\in\mathbb{R}^{D}$ to the action expert's sequence.
In every block $l$, we apply a directional attention mask $M$:
\begin{equation}
  M_{uv}=\begin{cases}-\infty & u\in V\cup A,\;v=J,\\ 0 & \text{otherwise,}\end{cases}
  \label{eq:mask}
\end{equation}
where $u$ indexes queries and $v$ indexes keys, allowing $J$ to attend to all visual, final-step action, and its own token states ($V\cup A\cup\{J\}$) while blocking video and action tokens from attending to $J$.

\paragraph{Token-gated adaptation.}
To extract value-relevant features without modifying frozen policy weights, for each projection $W\in\mathbb{R}^{d_{\mathrm{out}}\times d_{\mathrm{in}}}$ in $\{Q_l,K_l,W^V_l,O_l,W^{\mathrm{up}}_l,W^{\mathrm{down}}_l\}$ of the action expert, we attach rank-$r$ low-rank adapters~\cite{lora} ($A\in\mathbb{R}^{r\times d_{\mathrm{in}}}$, $B\in\mathbb{R}^{d_{\mathrm{out}}\times r}$) gated solely at $J$'s index $p_J$:
\begin{equation}
  W x_u \;\longmapsto\; W x_u+\1[u{=}p_J]\,\tfrac{\alpha}{r}\,B A x_u.
  \label{eq:gate}
\end{equation}
Evaluated as a single-query read branch over frozen $[V^{(l)};A^{(l)}]$ keys/values and $J$'s own adapted state, original tokens traverse the unaltered graph, keeping action predictions bit-identical.
After $L$ layers, a two-layer MLP $g_\theta$ maps $j^{(L)}$ to a scalar success probability:
\begin{equation}
  q_\theta(o_k,a_k)=\sigma\big(g_\theta(j^{(L)})\big),\qquad \theta=\{j^{(0)},A,B,g\}.
  \label{eq:readout}
\end{equation}
Because $J$ runs only in the final denoising step and reuses cached visual key--value states alongside frozen action-expert weights, our judge introduces only $13$M trainable parameters ($r{=}16$ on Fast-WAM) and adds $1.6$\,ms per chunk on an NVIDIA A800 GPU.

\paragraph{Proposed cross-fitted temporal-difference supervision.}
We deploy $\pi_{\mathrm{base}}$ on rollout scenes disjoint from evaluation to collect trajectories $\xi=(o_0,a_0,\dots,o_{T-1},a_{T-1})$ with binary outcomes $Y\in\{0,1\}$.
To propagate multi-step value estimates without self-bootstrapping bias---where bootstrapping on the same scenes propagates overfitting and degrades calibration---we propose cross-fitted TD training: we split rollout scenes into $K{=}2$ disjoint folds $(\mathcal{D}_{\mathrm{judge}}^{(1)},\mathcal{D}_{\mathrm{judge}}^{(2)})$, fit a self-bootstrapped $\mathrm{TD}(\lambda)$ predictor $\bar\theta^{(f)}$ per fold $f\in\{1,2\}$ (warmed up from Monte-Carlo outcomes $Y$), and compute out-of-fold targets $G_k^{(\lambda)}$ ($0\le k<T$) for $\xi\in\mathcal{D}_{\mathrm{judge}}^{(f)}$ via the complement predictor $\bar\theta^{(-f)}$:
\begin{equation}
  G_k^{(\lambda)}=(1{-}\lambda)\sum_{n=1}^{T-k-1}\lambda^{n-1}q_{\bar\theta^{(-f)}}(o_{k+n},a_{k+n})+\lambda^{T-k-1}Y.
  \label{eq:td_target}
\end{equation}
We then train a unified predictor $\theta$ across all folds by minimizing binary cross-entropy against $G_k^{(\lambda)}$:
\begin{equation}
  \begin{aligned}
    &\mathcal{L}_{J}(\theta)=-\frac{1}{\E_{\pi_{\mathrm{base}}}[|\xi|]}\E_{\xi\sim\pi_{\mathrm{base}}}\sum_{k=0}^{|\xi|-1}\!\Big[ G_k^{(\lambda)}\log q_\theta(o_k,a_k) \\[-3pt]
    &\qquad\qquad +(1{-}G_k^{(\lambda)})\log(1{-}q_\theta(o_k,a_k))\Big].
  \end{aligned}
  \label{eq:bce}
\end{equation}
Because binary log-loss is a strictly proper scoring rule, for any target generator its population minimizer recovers $\E_{\pi_{\mathrm{base}}}[G_k^{(\lambda)}\mid o_k,a_k]$.
Under the Markov property (proof in \cref{sec:supp:calib}), applying the tower property to the $\mathrm{TD}(\lambda)$ evaluation operator shows that for any $\lambda\in[0,1]$, its unique population fixed point is the true conditional success probability:
\begin{equation}
  q^\star(o_k,a_k)=\E_{\pi_{\mathrm{base}}}\big[Y\mid o_k,a_k\big].
  \label{eq:qstar}
\end{equation}
Meanwhile, evaluating $G_k^{(\lambda)}$ out-of-fold via $\bar\theta^{(-f)}$ eliminates same-scene self-bootstrapping bias in finite samples, enabling $q_\theta$ to learn well-calibrated success probabilities using only binary rollout success/failure signals---without human labels, dense rewards, or external detectors.

\subsection{Adaptive System 1 \& 2 Takeover}
\label{sec:method:defer}
\paragraph{Bidirectional hand-off and cost--performance control.}
At step $k$, the WAM samples a candidate chunk $a_k^{\mathrm{base}}\sim\pi_{\mathrm{base}}(\cdot\mid o_k)$ and evaluates $q_\theta(o_k,a_k^{\mathrm{base}})$ to schedule hand-offs with a GPT-6-Astra-based agent (\emph{Astra}, System~2).
With entry threshold $\tau$ and hysteresis threshold $\tau_{\mathrm{back}}\ge\tau$ ($\tau_k{=}\tau$ in WAM mode and $\tau_k{=}\tau_{\mathrm{back}}$ while Astra is active), control switches via:
\begin{equation}
  a_k=\left\{\!\begin{array}{@{}l@{\;\;}l@{}}a_k^{\mathrm{base}}, & \text{if } q_\theta(o_k,a_k^{\mathrm{base}})\ge\tau_k,\\[2pt] \pi_{\mathrm{Astra}}(o_k), & \text{if } q_\theta(o_k,a_k^{\mathrm{base}})<\tau_k.\end{array}\right.
  \label{eq:kappa}
\end{equation}
When invoked, Astra estimates 3D object poses from \mbox{RGB-D} observations and executes modular manipulation actions (grasping, re-grasping, placing, sweeping) from the live state left by the WAM.
After each action segment, the WAM proposes its next chunk and re-evaluates $q_\theta$; once $q_\theta\ge\tau_{\mathrm{back}}$, control returns seamlessly to $\pi_{\mathrm{base}}$ without resets.
Because $q_\theta$ is calibrated in probability units, $\tau$ directly sets the minimum acceptable success probability, smoothly trading System~2 call budget for task success.

\subsection{Consolidating Capability Gaps into System 1}
\label{sec:method:distill}
Because Astra intervenes only when $q_\theta<\tau_k$, its takeover trajectories pinpoint the exact states where the WAM lacks capability---charting its \emph{capability gaps}.
Successful System~1 \& 2 rollouts ($Y{=}1$, collected without human action labels) yield two complementary sources: (i)~$\mathcal{D}_{\mathrm{help}}$, Astra's recovery segments sliced into horizon-$H$ chunks, and (ii)~$\mathcal{D}_{\mathrm{own}}$, trajectories completed by the WAM alone.
Keeping the video expert and original $\pi_{\mathrm{base}}$ frozen, we fine-tune a separate repaired action expert $\phi_{\mathrm{rep}}$ (initialized from $\phi_{\mathrm{base}}$) via the standard unweighted flow-matching loss \eqref{eq:fm} on a mixed dataset $\mathcal{D}_{\mathrm{mix}}$:
\begin{equation}
  \phi_{\mathrm{rep}}=\arg\min_\phi\ \mathcal{L}_{\mathrm{FM}}\big(\phi;\,\mathcal{D}_{\mathrm{mix}}\big),
  \label{eq:mix}
\end{equation}
where each batch in $\mathcal{D}_{\mathrm{mix}}$ samples $(\mathcal{D}_{\mathrm{help}},\mathcal{D}_{\mathrm{own}},\mathcal{D}_{\mathrm{replay}})$ in fixed data proportions (\eg, $1{:}1{:}2$), pairing $\mathcal{D}_{\mathrm{help}}$ with mastered-skill anchor data ($\mathcal{D}_{\mathrm{own}}$ and replayed pre-training demonstrations $\mathcal{D}_{\mathrm{replay}}$): $\mathcal{D}_{\mathrm{help}}$ repairs bottleneck states, $\mathcal{D}_{\mathrm{own}}$ anchors native transitions, and $\mathcal{D}_{\mathrm{replay}}$ rehearses prior task skills into fast RGB-only weights.

\begin{table*}[t]
\centering
\caption{\textbf{Main benchmark results on RoboTwin~2.0 and LIBERO-Plus.}
We evaluate Fast-WAM~\cite{fastwam} and LingBot-VA~\cite{lingbotva} on full RoboTwin~2.0~\cite{robotwin2} (50 tasks; \emph{clean-only} training, reporting the mean across \emph{Clean} and \emph{Random} conditions~\cite{robotwin2,yang20264dwam}) and full LIBERO-Plus~\cite{liberoplus} (all suites across seven perturbation families~\cite{roboharness}), reporting success (\%) on \emph{Hard} ($<40\%$ base), \emph{Mastered} ($\ge75\%$ base), and \emph{All} evaluation conditions under a matched $30\%$ agent call budget.}
\label{tab:main}
\footnotesize
\setlength{\tabcolsep}{3.5pt}
\resizebox{\textwidth}{!}{%
\begin{tabular}{l c ccc ccc ccc ccc}
\toprule
& & \multicolumn{6}{c}{\textbf{Fast-WAM}~\cite{fastwam}} & \multicolumn{6}{c}{\textbf{LingBot-VA}~\cite{lingbotva}}\\
\cmidrule(lr){3-8}\cmidrule(lr){9-14}
& & \multicolumn{3}{c}{RoboTwin~2.0 (Full)} & \multicolumn{3}{c}{LIBERO-Plus (Full)} & \multicolumn{3}{c}{RoboTwin~2.0 (Full)} & \multicolumn{3}{c}{LIBERO-Plus (Full)}\\
\cmidrule(lr){3-5}\cmidrule(lr){6-8}\cmidrule(lr){9-11}\cmidrule(lr){12-14}
Method & Agent\% & Hard & Mast. & All & Hard & Mast. & All & Hard & Mast. & All & Hard & Mast. & All\\
\midrule
Base WAM & 0\% & 3.1 & 90.8 & 39.8 & 22.2 & 86.9 & 51.5 & 19.8 & 88.9 & 57.7 & 29.6 & 91.0 & 69.5\\
\midrule
\multicolumn{14}{l}{\emph{Two-System Collaborative Frameworks}}\\
Galbot-Astra~\cite{galbot2026astra} & 30\% & 38.4 & 91.7 & 60.9 & 48.6 & 90.4 & 68.7 & 49.6 & 92.6 & 74.3 & 52.8 & 93.4 & 79.6\\
Spotter~\cite{spotter} & 30\% & 41.9 & 91.3 & 62.8 & 55.4 & 89.8 & 72.9 & 51.2 & 91.4 & 73.8 & 61.2 & 92.7 & 82.4\\
CALL-Astra~\cite{callastra} & 30\% & 45.3 & 92.1 & 65.1 & 52.1 & 91.2 & 71.1 & 56.4 & 92.9 & 76.9 & 58.6 & 93.1 & 81.5\\
\rowcolor{oursblue!8}\textbf{\ours System 1 \& 2 (Ours)} & 30\% & \best{60.2} & \best{93.6} & \best{74.5} & \best{68.9} & \best{93.6} & \best{81.4} & \best{70.4} & \best{94.7} & \best{84.1} & \best{74.8} & \best{95.1} & \best{88.3}\\
\textcolor{gray}{GPT-6-Astra Agent (Upper Bound)} & \textcolor{gray}{100\%} & \textcolor{gray}{82.4} & \textcolor{gray}{96.5} & \textcolor{gray}{88.4} & \textcolor{gray}{84.2} & \textcolor{gray}{97.0} & \textcolor{gray}{90.6} & \textcolor{gray}{79.8} & \textcolor{gray}{95.9} & \textcolor{gray}{88.4} & \textcolor{gray}{81.6} & \textcolor{gray}{95.2} & \textcolor{gray}{90.6}\\
\midrule
\multicolumn{14}{l}{\emph{Consolidated Standalone WAM Policy}}\\
\rowcolor{oursblue!8}\textbf{\ours Standalone (Ours, Consolidated)} & 0\% & \best{42.4} & 90.4 & \best{63.4} & \best{54.8} & 86.3 & \best{71.6} & \best{55.6} & 88.7 & \best{75.5} & \best{61.6} & 91.2 & \best{82.6}\\
\bottomrule
\end{tabular}%
}
\end{table*}

\subsection{Erosion-Free Deployment via Dynamic Confidence Gating}
\label{sec:method:gate}
Fine-tuning on distribution-shifted interventions induces a fundamental repair--erosion trade-off.
Specifically, let $p{=}q_{\mathrm{base}}(o_0,a_0^{\mathrm{base}})$ denote calibrated initial confidence ($\E[Y_{\mathrm{base}}\mid p]{=}p$).
Under anchor retention $\alpha\in[0,1)$ on native behavior and sampling fraction $1{-}\alpha$ on intervention data whose standalone rollout capability is $c\in(0,1)$, expected performance change follows a \emph{competence shrinkage law} (proof and multi-step routing analysis in \cref{sec:analysis}):
\begin{equation}
  \Delta(p)\!\coloneqq\!\E\big[Y_{\mathrm{rep}}{-}Y_{\mathrm{base}}\mid q_{\mathrm{base}}{=}p\big]\!=(1{-}\alpha)(c{-}p).
  \label{eq:shrink_main}
\end{equation}
Wholesale replacement with $\pi_{\mathrm{rep}}$ thus \emph{repairs} low-confidence states ($p{<}c$, $\Delta(p){>}0$) by $(1{-}\alpha)\E[(c{-}q_{\mathrm{base}})_+]$ but \emph{degrades} mastered states ($p{>}c$, $\Delta(p){<}0$) by $(1{-}\alpha)\E[(q_{\mathrm{base}}{-}c)_+]$, whereas gating $\pi_{\mathrm{rep}}$ on $q_{\mathrm{base}}{<}\tau$ ($\tau{\approx}c$) activates repair only where $\Delta(p){>}0$.
Accordingly, we retain the frozen base judge $q_{\mathrm{base}}\coloneqq q_{\theta_{\mathrm{base}}}$ to gate each sampled chunk $a_k^{\mathrm{base}}\sim\pi_{\mathrm{base}}(\cdot\mid o_k)$ ($\tau_k{=}\tau$ in base mode; $\tau_k{=}\tau_{\mathrm{back}}$ while routing through $\pi_{\mathrm{rep}}$):
\begin{equation}
  a_k=\left\{\!\begin{array}{@{}l@{\;\;}l@{}}a_k^{\mathrm{base}}, & \text{if } q_{\mathrm{base}}(o_k,a_k^{\mathrm{base}})\ge\tau_k,\\[2pt] a_k^{\mathrm{rep}}\!\sim\!\pi_{\mathrm{rep}}(\cdot\!\mid\!o_k), & \text{if } q_{\mathrm{base}}(o_k,a_k^{\mathrm{base}})<\tau_k.\end{array}\right.
  \label{eq:switch}
\end{equation}
When $q_{\mathrm{base}}\ge\tau_k$, the robot executes $a_k^{\mathrm{base}}$ directly without resampling; when $q_{\mathrm{base}}<\tau_k$, it executes $a_k^{\mathrm{rep}}\sim\pi_{\mathrm{rep}}(\cdot\mid o_k)$ until $q_{\mathrm{base}}\ge\tau_{\mathrm{back}}$.
Sharing the frozen 5B video backbone caches visual KV once per observation, adds only $1.7$--$2.0$\,GB BF16 VRAM, and runs $\pi_{\mathrm{rep}}$ only on low-$q_{\mathrm{base}}$ chunks.

\subsection{Continual Self-Evolution Loop}
\label{sec:method:loop}
This judge--schedule--consolidate pipeline repeats over rounds $m{=}0,1,2,\dots$ for continual self-evolution.
Across all rounds, \textbf{$\pi_{\mathrm{base}}$ and $q_{\mathrm{base}}$ remain permanently frozen---the only policy updated is the repaired expert $\pi_{\mathrm{rep}}^{(m)}$}.
At round $m$ ($\pi^{(0)}{=}\pi_{\mathrm{base}}$), we treat $(\pi_{\mathrm{base}},\pi_{\mathrm{rep}}^{(m)})$ routed internally by $q_{\mathrm{base}}$ via \cref{eq:switch} \textbf{as one unified system $\pi^{(m)}$}, avoiding redundant Astra calls on already-repaired states in two steps:
\begin{enumerate}\setlength{\itemsep}{1.5pt}\setlength{\topsep}{2pt}
  \item \textbf{Judge the current unified system $\pi^{(m)}$:} On rollouts of $\pi^{(m)}$, we likewise use cross-fitted TD learning (\cref{sec:method:token}) to train (warm-started from the prior round, with $\theta^{(0)}{=}\theta_{\mathrm{base}}$) a system-level judge $q_{\theta^{(m)}}$ predicting whether $\pi^{(m)}$ as a whole will succeed from its output chunk $a_k$.
  \item \textbf{Collect residual gaps and update only $\pi_{\mathrm{rep}}^{(m+1)}$:} Running $\pi^{(m)}$ with Astra, we invoke Astra only when $q_{\theta^{(m)}}(o_k,a_k)<\tau_k$ (\ie, residual bottlenecks unsolved by both $\pi_{\mathrm{base}}$ and $\pi_{\mathrm{rep}}^{(m)}$); aggregating new segments with prior data~\cite{dagger}, we optimize \eqref{eq:mix} to \textbf{update only $\pi_{\mathrm{rep}}^{(m+1)}$}, pairing it with unchanged $(\pi_{\mathrm{base}},q_{\mathrm{base}})$ to form $\pi^{(m+1)}$.
\end{enumerate}
Thus, unchanged $(\pi_{\mathrm{base}},q_{\mathrm{base}})$ permanently safeguards prior mastery while $\pi_{\mathrm{rep}}^{(m)}$ steadily absorbs residual skills.

\section{Experiments}
\label{sec:exp}

\begin{figure*}[!t]
  \centering
  \includegraphics[width=\textwidth]{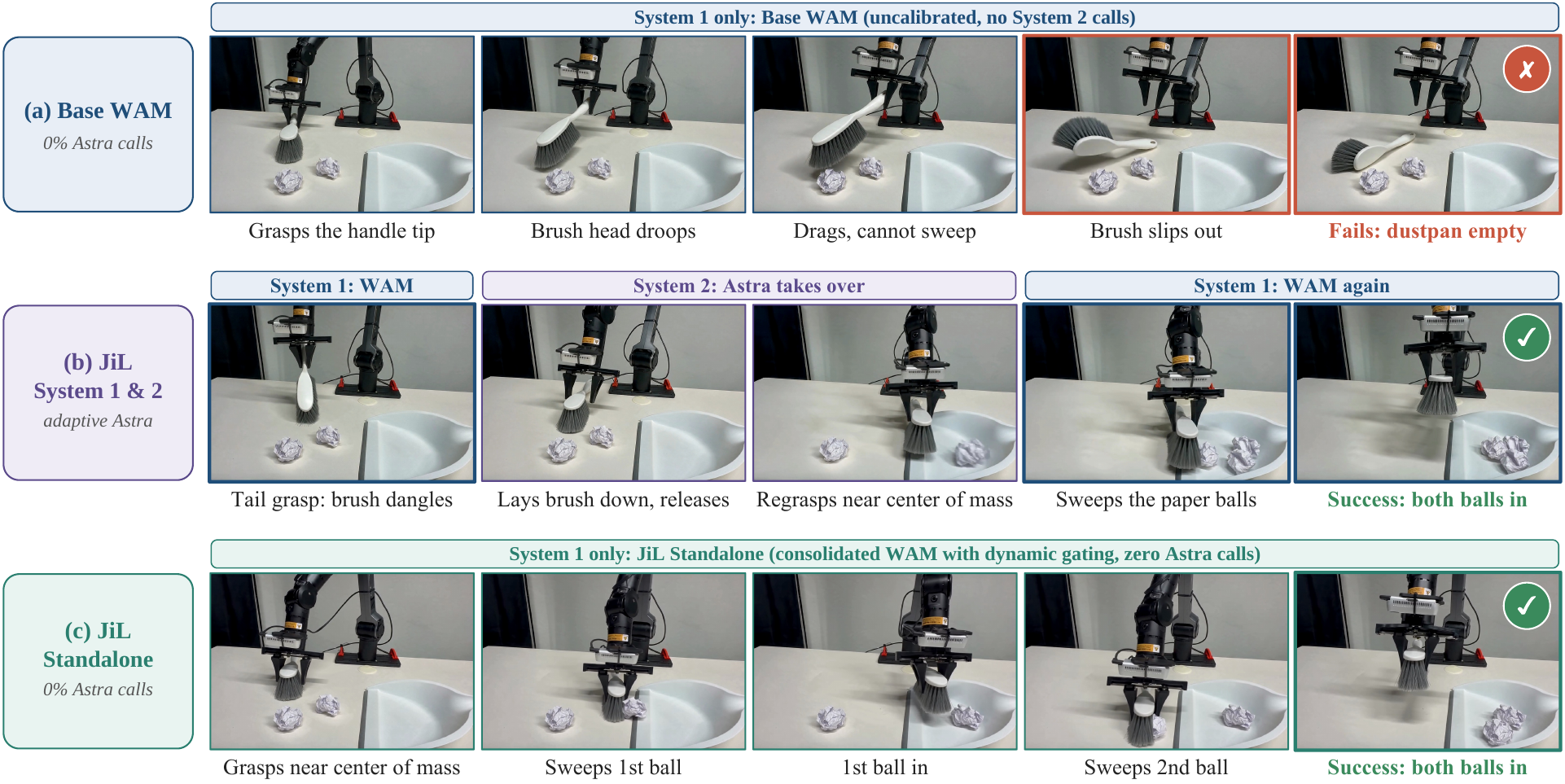}
  \caption{\textbf{Real-world robotic manipulation rollouts on \emph{Sweep Trash}.}
  Qualitative execution on physical hardware comparing \textbf{(a)~Base WAM}, where an unstable handle-tip grasp causes the brush to droop and slip out; \textbf{(b)~\ours System~1 \& 2}, where low confidence triggers Astra to lay the brush down, regrasp near the center of mass, and hand control back to the WAM to sweep both paper balls in; and \textbf{(c)~\ours Standalone} (consolidated WAM, zero test-time Astra calls), which directly grasps near the center of mass and sweeps both paper balls in.}
  \label{fig:real_demo}
\end{figure*}

\begin{table}[!t]
\centering
\caption{\textbf{Real-world robot manipulation results.}
Success rates (\%) across three physical tasks (50 trials per task) on Fast-WAM and LingBot-VA under $0\%$ standalone and matched $30\%$ agent call budgets.}
\label{tab:real}
\footnotesize
\setlength{\tabcolsep}{2.0pt}
\resizebox{\columnwidth}{!}{%
\begin{tabular}{l c cccc}
\toprule
Method & Agent\% & Sweep & Organize & Stack & \textbf{Avg.}\\
\midrule
\multicolumn{6}{l}{\emph{Backbone: Fast-WAM}~\cite{fastwam}}\\
Base WAM & 0\% & 42.0 & 38.0 & 52.0 & 44.0\\
\rowcolor{oursblue!8}\textbf{\ours Standalone (Ours)} & 0\% & \best{70.0} & \best{64.0} & \best{72.0} & \best{68.7}\\
\cmidrule(lr){1-6}
Galbot-Astra~\cite{galbot2026astra} & 30\% & 64.0 & 66.0 & 68.0 & 66.0\\
Spotter~\cite{spotter} & 30\% & 70.0 & 62.0 & 70.0 & 67.3\\
CALL-Astra~\cite{callastra} & 30\% & 66.0 & 68.0 & 76.0 & 70.0\\
\rowcolor{oursblue!8}\textbf{\ours System 1 \& 2 (Ours)} & 30\% & \best{84.0} & \best{76.0} & \best{86.0} & \best{82.0}\\
\midrule
\multicolumn{6}{l}{\emph{Backbone: LingBot-VA}~\cite{lingbotva}}\\
Base WAM & 0\% & 48.0 & 46.0 & 58.0 & 50.7\\
\rowcolor{oursblue!8}\textbf{\ours Standalone (Ours)} & 0\% & \best{74.0} & \best{70.0} & \best{76.0} & \best{73.3}\\
\cmidrule(lr){1-6}
Galbot-Astra~\cite{galbot2026astra} & 30\% & 70.0 & 72.0 & 72.0 & 71.3\\
Spotter~\cite{spotter} & 30\% & 76.0 & 68.0 & 76.0 & 73.3\\
CALL-Astra~\cite{callastra} & 30\% & 70.0 & 70.0 & 78.0 & 72.7\\
\rowcolor{oursblue!8}\textbf{\ours System 1 \& 2 (Ours)} & 30\% & \best{86.0} & \best{82.0} & \best{90.0} & \best{86.0}\\
\midrule
\textcolor{gray}{GPT-6-Astra (Upper Bound)} & \textcolor{gray}{100\%} & \textcolor{gray}{90.0} & \textcolor{gray}{86.0} & \textcolor{gray}{92.0} & \textcolor{gray}{89.3}\\
\bottomrule
\end{tabular}%
}
\end{table}

\subsection{Experimental Setup}
\label{sec:exp:setup}

\paragraph{Models, benchmarks, and stratification.}
We instantiate \ours on three WAM backbones spanning distinct architectures and test-time paradigms: \textbf{Fast-WAM}~\cite{fastwam} (skipping test-time video decoding), \textbf{LingBot-VA}~\cite{lingbotva} (autoregressive video--action MoT), and \textbf{Kairos}~\cite{kairos} (a native unified VideoDiT--ActionDiT MoT equipped with hybrid linear temporal attention).
Following public protocols~\cite{robotwin2,yang20264dwam,robotwinplus,roboharness}, we evaluate Fast-WAM and LingBot-VA on:
(i)~\textbf{RoboTwin~2.0}~\cite{robotwin2} (50 bimanual tasks; \emph{clean-only} training across all 100 \emph{Clean} and \emph{Random} conditions~\cite{robotwin2,yang20264dwam}); and
(ii)~\textbf{LIBERO-Plus}~\cite{liberoplus} (all 28 suite--perturbation conditions across seven perturbation families~\cite{roboharness}), stratified by base success into \emph{Hard} ($<40\%$), \emph{Mastered} ($\ge75\%$), and \emph{All}.
To verify generality on a stronger native WAM backbone, we further evaluate \ours on \textbf{Kairos} under the same full RoboTwin~2.0 \emph{clean-only} protocol.
On a 6-DoF AgileX Piper arm with a RealSense D435 camera, we evaluate Fast-WAM and LingBot-VA on three physical tasks (50 trials each): \textbf{Sweep Trash}, \textbf{Organize Table}, and \textbf{Stack Blocks}.

\paragraph{Baselines, budget matching, and disjoint seeds.}
In \cref{tab:main,tab:real,tab:efficiency}, all two-system methods share the same base WAM, \mbox{RGB-D} GPT-6-Astra-based agent, and $30\%$ agent call budget (measured at the chunk level), and \cref{tab:kairos} evaluates \ours on Kairos under the identical \mbox{RGB-D} agent and $30\%$ budget:
(i)~\textbf{Galbot-Astra}~\cite{galbot2026astra} (periodic $30\%$ chunk review/override);
(ii)~\textbf{Spotter}~\cite{spotter} (local VLM screener triggering primitive Astra repairs); and
(iii)~\textbf{CALL-Astra}~\cite{callastra} (counterfactual MLP head on frozen policy features for windowed hand-offs).
\ours uses rank-16 token-gated LoRA ($\alpha{=}32$), 2-fold cross-fitted $\mathrm{TD}(\lambda{=}0.8)$, and data sampling ratio $1{:}1{:}2$ (\cref{eq:mix}).
Thresholds are calibrated on held-out $\mathcal{D}_{\mathrm{cal}}$ to match target budgets ($\tau_{\mathrm{back}}{-}\tau{=}0.05$).
Each consolidation round collects 50 label-free System~1 \& 2 rollouts per task in simulation (25 per physical task; 1 round except \cref{fig:repair}b).
Judge training, $\mathcal{D}_{\mathrm{cal}}$, consolidation ($\mathcal{D}_{\mathrm{help}},\mathcal{D}_{\mathrm{own}}$), and test evaluation use \textbf{four strictly disjoint scene-seed sets} on one NVIDIA A800 GPU.

\begin{table}[!t]
\centering
\caption{\textbf{Results on Kairos~\cite{kairos} (RoboTwin~2.0).}
Success rates (\%) on \emph{Hard}, \emph{Mastered}, and \emph{All} conditions under $0\%$ standalone and $30\%$ agent budgets; gains over Base Kairos are shown in parentheses.}
\label{tab:kairos}
\footnotesize
\setlength{\tabcolsep}{3.0pt}
\resizebox{\columnwidth}{!}{%
\begin{tabular}{l c ccc}
\toprule
Method & Agent\% & Hard & Mastered & \textbf{All}\\
\midrule
Base Kairos~\cite{kairos} & 0\% & 17.7 & 92.2 & 77.3\\
\rowcolor{oursblue!8}\textbf{\ours Standalone (Ours)} & 0\% & \best{51.9}\,{\scriptsize\textcolor{oursblue}{(+34.2)}} & \best{94.1}\,{\scriptsize\textcolor{oursblue}{(+1.9)}} & \best{83.5}\,{\scriptsize\textcolor{oursblue}{(+6.2)}}\\
\cmidrule(lr){1-5}
\rowcolor{oursblue!8}\textbf{\ours System 1 \& 2 (Ours)} & 30\% & \best{77.1}\,{\scriptsize\textcolor{oursblue}{(+59.4)}} & \best{95.8}\,{\scriptsize\textcolor{oursblue}{(+3.6)}} & \best{89.8}\,{\scriptsize\textcolor{oursblue}{(+12.5)}}\\
\bottomrule
\end{tabular}%
}
\end{table}

\subsection{Main Results}
\label{sec:exp:main}

\paragraph{Simulation and real-world benchmarks.}
\Cref{tab:main,tab:real} summarize simulation and physical results.
Because calibrated pre-execution judgment focuses the $30\%$ Astra budget on impending bottlenecks rather than confident states or post-drift recovery, \textbf{\ours System~1 \& 2} outperforms all baselines in every setting---boosting \emph{Hard} success by $+13.5\%$ to $+14.9\%$ over the best baseline and lifting overall success to $74.5\%$ and $81.4\%$ on Fast-WAM ($+9.4\%$ and $+8.5\%$) and $84.1\%$ and $88.3\%$ on LingBot-VA ($+7.2\%$ and $+5.9\%$), nearing the $100\%$ Astra ceiling ($88.4\%$--$90.6\%$).
With one consolidation round and \textbf{zero test-time agent calls}, \textbf{\ours Standalone} reaches $63.4\%$ and $71.6\%$ on Fast-WAM ($+23.6\%$ and $+20.1\%$) and $75.5\%$ and $82.6\%$ on LingBot-VA ($+17.8\%$ and $+13.1\%$), matching $30\%$-budget two-system baselines while keeping mastered skills within $0.6\%$ of the base WAM (\cref{tab:main}).
On physical hardware (\cref{tab:real}), \ours System~1 \& 2 reaches $82.0\%$ and $86.0\%$, while \ours Standalone---with \textbf{no agent calls at test time}---achieves $68.7\%$ ($+24.7\%$) and $73.3\%$ ($+22.6\%$).

\paragraph{Scaling to a stronger native WAM (Kairos).}
\Cref{tab:kairos} further evaluates \ours on \textbf{Kairos}~\cite{kairos} under the same RoboTwin~2.0 protocol.
Even on this stronger base model ($77.3\%$ overall), \textbf{\ours System~1 \& 2} at a $30\%$ budget lifts \emph{Hard} success by $+59.4\%$ ($17.7\%{\to}77.1\%$) and overall success to $89.8\%$ ($+12.5\%$), surpassing even the $100\%$ pure GPT-6-Astra agent ($88.4\%$ in \cref{tab:main}).
With \textbf{zero test-time agent calls}, \textbf{\ours Standalone} improves \emph{Hard} by $+34.2\%$ ($51.9\%$) and overall success to $83.5\%$ ($+6.2\%$) while also lifting \emph{Mastered} performance ($92.2\%{\to}94.1\%$).

\begin{figure}[!t]
  \centering
  \includegraphics[width=\linewidth]{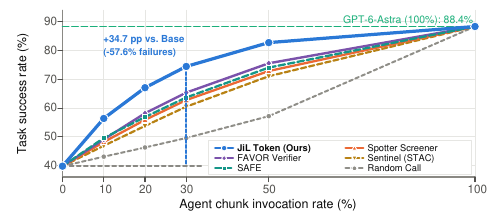}
  \caption{\textbf{System~1 \& 2 success across agent call budgets} (Fast-WAM on RoboTwin~2.0), plugging our judgment token vs.\ Spotter Screener~\cite{spotter}, Sentinel (STAC)~\cite{sentinel}, FAVOR Verifier~\cite{favor}, SAFE~\cite{safe}, and Random Invocation into the same pipeline.}
  \label{fig:calib_sys12}
\end{figure}

\begin{table}[!t]
\centering
\caption{\textbf{Self-assessment accuracy, calibration, and latency} on Fast-WAM (RoboTwin~2.0, NVIDIA A800): same-state pair accuracy ($\mathrm{PairAcc}_{a}$, $1{,}000$ pairs), trigger precision ($\mathrm{Prec}_{\mathrm{trig}}$), failure miss rate ($\mathrm{Miss}_{\mathrm{fail}}$), expected calibration error ($\mathrm{ECE}$, $5{,}000$ rollouts), and per-chunk latency ($\Delta t$).}
\label{tab:calibmain}
\footnotesize
\setlength{\tabcolsep}{1.4pt}
\resizebox{\columnwidth}{!}{%
\begin{tabular}{l ccccc}
\toprule
Method & $\mathrm{PairAcc}_{a}\uparrow$ & $\mathrm{Prec}_{\mathrm{trig}}\uparrow$ & $\mathrm{Miss}_{\mathrm{fail}}\downarrow$ & $\mathrm{ECE}\downarrow$ & $\Delta t$\,(ms)$\downarrow$\\
\midrule
Spotter Screener~\cite{spotter} & 69.5\% & 86.6\% & 38.1\% & 24.8\% & 168.4\\
Sentinel (STAC)~\cite{sentinel} & 65.4\% & 84.7\% & 42.4\% & 27.5\% & 412.0\\
FAVOR Verifier~\cite{favor} & 71.8\% & 87.2\% & 30.9\% & 19.2\% & 96.5\\
SAFE~\cite{safe} & 74.3\% & 86.2\% & 33.6\% & 16.4\% & 2.1\\
\rowcolor{oursblue!8}\textbf{Judgment Token (Ours)} & \best{86.7\%} & \best{89.9\%} & \best{13.8\%} & \best{3.1\%} & \best{1.6}\\
\midrule
\multicolumn{6}{l}{\emph{Ablation Studies}}\\
\quad w/o Token-Gated LoRA & 58.3\% & 73.7\% & 70.3\% & 14.2\% & 1.0\\
\quad Naive TD (w/o Cross-Fit) & 81.9\% & 87.9\% & 22.8\% & 8.7\% & 1.6\\
\quad Monte-Carlo ($\lambda{=}1.0$) & 72.4\% & 85.4\% & 36.2\% & 6.8\% & 1.6\\
\bottomrule
\end{tabular}%
}
\end{table}

\subsection{Self-Assessment, Scheduling \& Efficiency}
\label{sec:exp:calib}

\paragraph{Judging action quality and calibration vs.\ baselines.}
We evaluate Fast-WAM on RoboTwin~2.0 over two held-out splits (metric definitions in \cref{sec:supp:metrics}):
(i)~$1{,}000$ counterfactual same-state pairs $\mathcal{D}_{\mathrm{pair}}$ for $\mathrm{PairAcc}_a$---scoring our token and SAFE~\cite{safe} pre-execution while granting post-execution monitors (Spotter~\cite{spotter}, Sentinel~\cite{sentinel}, FAVOR~\cite{favor}) one executed chunk ($o_{k+1}$); and
(ii)~$5{,}000$ un-intervened $\pi_{\mathrm{base}}$ rollouts $\mathcal{D}_{\mathrm{eval}}$ for $\mathrm{Prec}_{\mathrm{trig}}$, $\mathrm{Miss}_{\mathrm{fail}}$, and episode-weighted 15-bin $\mathrm{ECE}$.
In \cref{tab:calibmain}, our judgment token outperforms all baselines on every metric at $1.6$\,ms per chunk on an NVIDIA A800 GPU, reaching $86.7\%$ $\mathrm{PairAcc}_a$ ($+12.4\%$ over the best baseline), $89.9\%$ $\mathrm{Prec}_{\mathrm{trig}}$, $13.8\%$ $\mathrm{Miss}_{\mathrm{fail}}$ (over $50\%$ fewer missed failures at $60\times$--$250\times$ lower latency than rollout/VLM/multi-sample monitors), and the lowest $\mathrm{ECE}$ of $3.1\%$ (vs.\ $16.4\%$--$27.5\%$).
In ablations, dropping token-gated LoRA drops $\mathrm{PairAcc}_a$ to $58.3\%$ ($70.3\%$ $\mathrm{Miss}_{\mathrm{fail}}$, $14.2\%$ $\mathrm{ECE}$) as frozen denoising projections cannot route cross-token failure cues into $j^{(0)}$.
Crucially, while Naive TD improves $\mathrm{PairAcc}_a$ over Monte-Carlo ($81.9\%$ vs.\ $72.4\%$), same-fold bootstrapping bias degrades its $\mathrm{ECE}$ to $8.7\%$ vs.\ unbiased MC's $6.8\%$; 2-fold cross-fitting eliminates this bias, achieving both $3.1\%$ $\mathrm{ECE}$ and $86.7\%$ $\mathrm{PairAcc}_a$.

\paragraph{Multi-budget scheduling and system efficiency.}
Across budgets from $10\%$ to $50\%$ (\cref{fig:calib_sys12}), our token consistently surpasses every baseline monitor, gaining $+34.7\%$ over Base WAM ($-57.6\%$ failures vs.\ $-16.3\%$ for random calls) at $30\%$.
In \cref{tab:efficiency}, \ours System~1 \& 2 cuts full-Astra episode time from $94.6$\,s to $31.4$\,s ($3.0\times$, beating Galbot-Astra's $33.8$\,s and Spotter's $43.2$\,s), while \ours Standalone runs in $17.3$\,s ($5.5\times$, $+0.5$\,s over Base WAM from $1.6$\,ms per-chunk judging and rare $\pi_{\mathrm{rep}}$ resampling).

\setcounter{topnumber}{2}
\begin{table}[!t]
\centering
\caption{\textbf{System efficiency comparison} on RoboTwin~2.0 (Fast-WAM, NVIDIA A800): agent call rate, per-chunk judge latency ($\Delta t_{\mathrm{judge}}$), mean episode time ($T_{\mathrm{ep}}$), speedup vs.\ full Astra, and success rate.}
\label{tab:efficiency}
\footnotesize
\setlength{\tabcolsep}{1.6pt}
\resizebox{\columnwidth}{!}{%
\begin{tabular}{l c ccc c}
\toprule
System / Method & Call\,(\%) & $\Delta t_{\mathrm{judge}}$\,(ms)$\downarrow$ & $T_{\mathrm{ep}}$\,(s)$\downarrow$ & Speed$\uparrow$ & Succ.\,(\%)$\uparrow$\\
\midrule
\multicolumn{6}{l}{\emph{Standalone Visuomotor Policies (0\% Agent Calls at Test Time)}}\\
Base WAM & 0 & -- & \best{16.8} & \best{$5.6\times$} & 39.8\\
\rowcolor{oursblue!8}\textbf{\ours Standalone} & 0 & 1.6 & 17.3 & $5.5\times$ & \best{63.4}\\
\midrule
\multicolumn{6}{l}{\emph{Two-System Collaborative Frameworks (30\% Agent Call Budget)}}\\
Galbot-Astra~\cite{galbot2026astra} & 30 & 0.0 & 33.8 & $2.8\times$ & 60.9\\
Spotter~\cite{spotter} & 30 & 168.4 & 43.2 & $2.2\times$ & 62.8\\
CALL-Astra~\cite{callastra} & 30 & 2.4 & 34.9 & $2.7\times$ & 65.1\\
\rowcolor{oursblue!8}\textbf{\ours System 1 \& 2} & 30 & 1.6 & \best{31.4} & \best{$3.0\times$} & \best{74.5}\\
\midrule
\textcolor{gray}{GPT-6-Astra (Full)} & \textcolor{gray}{100} & \textcolor{gray}{--} & \textcolor{gray}{94.6} & \textcolor{gray}{$1.0\times$} & \textcolor{gray}{88.4}\\
\bottomrule
\end{tabular}%
}
\end{table}

\begin{figure}[!t]
  \centering
  \includegraphics[width=\linewidth]{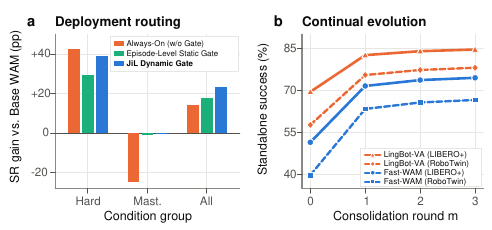}
  \caption{\textbf{Deployment routing and continual self-evolution.}
  \textbf{(a)}~Success-rate gain (\%) vs.\ Base WAM on Fast-WAM (RoboTwin~2.0, 1 round) across Hard, Mastered, and All conditions.
  \textbf{(b)}~Standalone success across consolidation rounds $m\in\{0,1,2,3\}$ ($m{=}0$: Base WAM) on RoboTwin~2.0 and LIBERO-Plus.}
  \label{fig:repair}
\end{figure}

\begin{table}[!t]
\centering
\caption{\textbf{Ablation of consolidation training data mixture} for $\pi_{\mathrm{rep}}$ on Fast-WAM (RoboTwin~2.0, 1 round), with dynamic confidence gating ($\pi_{\mathrm{base}}\leftrightarrow\pi_{\mathrm{rep}}$) enabled across all consolidated variants.}
\label{tab:distill}
\footnotesize
\setlength{\tabcolsep}{1.6pt}
\resizebox{\columnwidth}{!}{%
\begin{tabular}{ccc l ccc}
\toprule
$\mathcal{D}_{\mathrm{help}}$ & $\mathcal{D}_{\mathrm{own}}$ & $\mathcal{D}_{\mathrm{replay}}$ & Consolidation Data Mixture & Hard & Mast. & \textbf{All}\\
\midrule
-- & -- & -- & Base WAM & 3.1 & \best{90.8} & 39.8\\
-- & \cmark & \cmark & w/o $\mathcal{D}_{\mathrm{help}}$ (No Agent) & 3.6 & 90.5 & 40.0\\
\cmark & -- & -- & $\mathcal{D}_{\mathrm{help}}$ Only & 29.4 & 84.6 & 52.8\\
\cmark & -- & \cmark & w/o $\mathcal{D}_{\mathrm{own}}$ Anchoring & 36.2 & 88.1 & 58.6\\
\cmark & \cmark & -- & w/o $\mathcal{D}_{\mathrm{replay}}$ Rehearsal & 38.5 & 88.9 & 60.4\\
-- & \cmark & \cmark & Full-Astra Rollouts (repl.\ $\mathcal{D}_{\mathrm{help}}$) & 27.8 & 90.1 & 54.3\\
\rowcolor{oursblue!8}\cmark & \cmark & \cmark & \textbf{Full Mixture (\ours Standalone)} & \best{42.4} & 90.4 & \best{63.4}\\
\bottomrule
\end{tabular}%
}
\end{table}

\subsection{Capability-Gap Consolidation \& Evolution}
\label{sec:exp:distill}

\paragraph{Consolidation data mixture (\cref{tab:distill}).}
With dynamic gating active across all consolidated variants, \cref{tab:distill} ablates the components of $\mathcal{D}_{\mathrm{mix}}$ in \cref{eq:mix}.
Omitting Astra (\lq w/o $\mathcal{D}_{\mathrm{help}}$\rq) yields negligible change ($40.0\%$).
Training $\pi_{\mathrm{rep}}$ on $\mathcal{D}_{\mathrm{help}}$ alone lifts hard success to $29.4\%$ ($52.8\%$ overall) but drops mastered scenes to $84.6\%$ under occasional false-positive gate triggers; adding $\mathcal{D}_{\mathrm{replay}}$ ($58.6\%$) and $\mathcal{D}_{\mathrm{own}}$ anchors hand-off transitions, reaching $42.4\%$ hard, $90.4\%$ mastered, and $63.4\%$ overall.
Crucially, replacing $\mathcal{D}_{\mathrm{help}}$ with an equal chunk count of un-targeted full-episode Astra trajectories (\lq Full-Astra Rollouts\rq) drops from $42.4\%$ to $27.8\%$ ($54.3\%$ overall), as full rollouts spend budget on routine states while missing the exact off-manifold bottlenecks $\pi_{\mathrm{base}}$ enters---confirming that confidence-triggered interventions pinpoint true capability gaps.

\paragraph{Deployment gating and continual evolution (\cref{fig:repair}).}
\Cref{fig:repair}a fixes the full-mixture $\pi_{\mathrm{rep}}$ and ablates routing: consistent with \cref{eq:shrink_main}, `Always-On (w/o Gate)' gains $+42.7\%$ on hard conditions but suffers $-24.6\%$ mastered degradation ($+14.4\%$ overall, below `Episode-Level Static Gate' at $+18.0\%$).
Chunk-level dynamic gating ($\pi_{\mathrm{base}}\leftrightarrow\pi_{\mathrm{rep}}$) retains $92\%$ of hard repair ($+39.3\%$) with $-0.4\%$ mastered drift ($+23.6\%$ overall).
Across rounds $m{=}0{\to}1{\to}2{\to}3$ (\cref{fig:repair}b), round~1 captures the main gap jump while rounds~2--3 bring steady gains ($69.5\%{\to}82.6\%{\to}84.0\%{\to}84.6\%$ and $57.7\%{\to}75.5\%{\to}77.3\%{\to}78.1\%$ on LingBot-VA; $51.5\%{\to}71.6\%{\to}73.7\%{\to}74.5\%$ and $39.8\%{\to}63.4\%{\to}65.7\%{\to}66.6\%$ on Fast-WAM) toward the RGB-only observability ceiling and $q_{\mathrm{base}}$'s miss floor.

\section{Conclusion}
\label{sec:conclusion}

We presented \oursfull (\ours), a self-evolving System~1 \& 2 framework that calibrates world action models to recognize impending failures, progressing from adaptive System~2 takeover at runtime to System-2-free standalone deployment.
Trained via cross-fitted TD learning using only binary rollout success/failure signals, our causal judgment token predicts calibrated success probabilities in $1.6$\,ms per chunk on an NVIDIA A800 GPU without affecting action predictions, invoking a GPT-6-Astra-based agent on difficult states and reclaiming control once confidence recovers.
Guided by our competence shrinkage law, mastered-skill anchoring and dynamic confidence gating continually consolidate System~2's takeover trajectories into standalone WAMs---spanning Fast-WAM, LingBot-VA, and Kairos---across RoboTwin~2.0, LIBERO-Plus, and physical robot tasks without degrading existing capabilities.

{
    \small
    \renewcommand{\baselinestretch}{0.955}\selectfont
    \setlength{\bibsep}{0pt plus 0.2pt}
    \bibliographystyle{ieeenat_fullname}
    \bibliography{main}

\begin{thebibliography}{50}
\providecommand{\natexlab}[1]{#1}
\providecommand{\url}[1]{\texttt{#1}}
\expandafter\ifx\csname urlstyle\endcsname\relax
  \providecommand{\doi}[1]{doi: #1}\else
  \providecommand{\doi}{doi: \begingroup \urlstyle{rm}\Url}\fi

\bibitem[Agia et~al.(2024)Agia, Sinha, Yang, Cao, Antonova, Pavone, and
  Bohg]{sentinel}
Christopher Agia, Rohan Sinha, Jingyun Yang, Ziang Cao, Rika Antonova, Marco
  Pavone, and Jeannette Bohg.
\newblock {Unpacking Failure Modes of Generative Policies: Runtime Monitoring
  of Consistency and Progress}.
\newblock In \emph{CoRL}, 2024.

\bibitem[Angelopoulos and Bates(2023)]{conformal}
Anastasios~N. Angelopoulos and Stephen Bates.
\newblock {A Gentle Introduction to Conformal Prediction and Distribution-Free
  Uncertainty Quantification}.
\newblock \emph{Foundations and Trends in Machine Learning}, 16\penalty0
  (4):\penalty0 494--591, 2023.

\bibitem[Biderman et~al.(2024)Biderman, Portes, Ortiz, Paul, Greengard,
  Jennings, King, Havens, Chiley, Frankle, et~al.]{biderman2024lora}
Dan Biderman, Jacob Portes, Jose Javier~Gonzalez Ortiz, Mansheej Paul, Philip
  Greengard, Connor Jennings, Daniel King, Sam Havens, Vitaliy Chiley, Jonathan
  Frankle, et~al.
\newblock {LoRA Learns Less and Forgets Less}.
\newblock \emph{TMLR}, 2024.

\bibitem[Chen et~al.(2026)Chen, Xie, Fu, Yu, Pacini, Bajamahal, Kim, Drake,
  Kim, Xue, et~al.]{chen2026gap}
Kaiyuan Chen, Shuangyu Xie, Letian Fu, Justin Yu, William Pacini, Sandeep
  Bajamahal, Hudson Kim, Jaimyn Drake, Daehwa Kim, Haoru Xue, et~al.
\newblock {GaP: A Graph-as-Policy Multi-Agent Self-Learning Harness For
  Variational Automation Tasks}.
\newblock arXiv:2607.05369, 2026.

\bibitem[Chen et~al.(2025)Chen, Chen, Chen, Cai, Liu, Li, Liang, Lin, Ge, Gu,
  Deng, Guo, et~al.]{robotwin2}
Tianxing Chen, Zanxin Chen, Baijun Chen, Zijian Cai, Yibin Liu, Zixuan Li,
  Qiwei Liang, Xianliang Lin, Yiheng Ge, Zhenyu Gu, Weiliang Deng, Yubin Guo,
  et~al.
\newblock {RoboTwin 2.0: A Scalable Data Generator and Benchmark with Strong
  Domain Randomization for Robust Bimanual Robotic Manipulation}.
\newblock arXiv:2506.18088, 2025.

\bibitem[Cheng et~al.(2026)Cheng, Yi, Fang, Zhang, Feng, Li, Zhuang, Wang,
  Yang, Song, et~al.]{cheng2026gptpolicy}
Dongzhou Cheng, Taoran Yi, Ye Fang, Xingwu Zhang, Fan Feng, Yixuan Li,
  Gengxiong Zhuang, Rongze Wang, Shuai Yang, Wei Song, et~al.
\newblock {In-Context Robot Learning with VLM Agents}.
\newblock arXiv:2609.19138, 2026.

\bibitem[Du et~al.(2023)Du, Yang, Dai, Dai, Nachum, Tenenbaum, Schuurmans, and
  Abbeel]{unipi}
Yilun Du, Mengjiao Yang, Bo Dai, Hanjun Dai, Ofir Nachum, Joshua~B. Tenenbaum,
  Dale Schuurmans, and Pieter Abbeel.
\newblock {Learning Universal Policies via Text-Guided Video Generation}.
\newblock In \emph{NeurIPS}, 2023.

\bibitem[Fei et~al.(2025)Fei, Wang, Shi, Dai, Cai, Qian, Ji, He, Zhang, Fei,
  Fu, Gong, et~al.]{liberoplus}
Senyu Fei, Siyin Wang, Junhao Shi, Zihao Dai, Jikun Cai, Pengfang Qian, Li Ji,
  Xinzhe He, Shiduo Zhang, Zhaoye Fei, Jinlan Fu, Jingjing Gong, et~al.
\newblock {LIBERO-Plus: In-Depth Robustness Analysis of Vision-Language-Action
  Models}.
\newblock arXiv:2510.13626, 2025.

\bibitem[{Galbot Team} et~al.(2026){Galbot Team}, Chen, Cheng, Deng, Ding,
  Dong, Gao, Jia, Li, Li, Lian, et~al.]{galbot2026astra}
{Galbot Team}, Xuchuan Chen, Xiaoqian Cheng, Yu Deng, Lihe Ding, Shaocong Dong,
  Xiangjun Gao, Haozhe Jia, Zekai Li, Zhoujian Li, Yunrui Lian, et~al.
\newblock {Systematically Exploring the Capabilities of GPT-6 Astra as Embodied
  Policies}.
\newblock arXiv:2609.38537, 2026.

\bibitem[Grislain et~al.(2025)Grislain, Rahimi, Sigaud, and
  Chetouani]{grislain2026ifailsense}
Clemence Grislain, Hamed Rahimi, Olivier Sigaud, and Mohamed Chetouani.
\newblock {I-FailSense: Towards General Robotic Failure Detection with
  Vision-Language Models}.
\newblock arXiv:2509.16072, 2025.

\bibitem[Gu et~al.(2025)Gu, Ju, Sun, Gilitschenski, Nishimura, Itkina, and
  Shkurti]{safe}
Qiao Gu, Yuanliang Ju, Shengxiang Sun, Igor Gilitschenski, Haruki Nishimura,
  Masha Itkina, and Florian Shkurti.
\newblock {SAFE: Multitask Failure Detection for Vision-Language-Action
  Models}.
\newblock arXiv:2506.09937, 2025.

\bibitem[Guo et~al.(2026)Guo, Mo, Wang, Zhang, Wang, Deng, Zhang, Rao, and
  Hu]{guo2026robodawn}
Meng-Hao Guo, Zhe-Han Mo, Jia-Jun Wang, Yi Zhang, Kejin Wang, Yi-Xuan Deng,
  Jia-Peng Zhang, Yongming Rao, and Shi-Min Hu.
\newblock {Transferring the Intelligence of VLMs to Robotic Control}.
\newblock arXiv:2609.22966, 2026.

\bibitem[He et~al.(2026)He, Xie, Cao, Chen, Duan, Ge, Huo, Shao, and
  Tian]{he2026astra}
Sida He, Lingxi Xie, Yunning Cao, Pengfei Chen, Kaiwen Duan, Jiannan Ge, Xinyue
  Huo, Jiacheng Shao, and Qi Tian.
\newblock {Robot Manipulation with GPT-6-Astra: Body Knowledge, Experience
  Reuse, Emergent Skills, and Sim2Real Transfer}.
\newblock arXiv:2609.31770, 2026.

\bibitem[Hu et~al.(2022)Hu, Shen, Wallis, Allen-Zhu, Li, Wang, Wang, and
  Chen]{lora}
Edward~J. Hu, Yelong Shen, Phillip Wallis, Zeyuan Allen-Zhu, Yuanzhi Li, Shean
  Wang, Lu Wang, and Weizhu Chen.
\newblock {LoRA: Low-Rank Adaptation of Large Language Models}.
\newblock In \emph{ICLR}, 2022.

\bibitem[Huang et~al.(2026{\natexlab{a}})Huang, Gu, Zhang, Zou, Dong, Cen,
  Wang, and Zhang]{vlaconf}
Dehao Huang, Aoxiang Gu, Chengjie Zhang, Bolin Zou, Wenlong Dong, Zilang Cen,
  Yue Wang, and Hong Zhang.
\newblock {VLAConf: Calibrated Task-Success Confidence for
  Vision-Language-Action Models}.
\newblock arXiv:2605.29605, 2026{\natexlab{a}}.

\bibitem[Huang et~al.(2026{\natexlab{b}})Huang, Li, Hu, Qi, Xiao, Coates, Cao,
  Zhang, and Zhang]{roboharness}
Jinbang Huang, Zhiyuan Li, Yuanzhao Hu, Ran Qi, Yixin Xiao, Mark Coates,
  Tongtong Cao, Zhanguang Zhang, and Yingxue Zhang.
\newblock {RoboHarness: Memory-Driven Orchestration of Heterogeneous Robot
  Policies for Long-Horizon Planning}.
\newblock arXiv:2607.18060, 2026{\natexlab{b}}.

\bibitem[Huang et~al.(2024)Huang, Wang, Li, Zhang, and Fei-Fei]{rekep}
Wenlong Huang, Chen Wang, Yunzhu Li, Ruohan Zhang, and Li Fei-Fei.
\newblock {ReKep: Spatio-Temporal Reasoning of Relational Keypoint Constraints
  for Robotic Manipulation}.
\newblock In \emph{CoRL}, 2024.

\bibitem[Jia et~al.(2026)Jia, Lin, Zhang, Zhang, Liu, and Jiang]{jia2026agp}
Mengzhao Jia, Yang Lin, Xixin Zhang, Zhihan Zhang, Xiaobai Liu, and Meng Jiang.
\newblock {Agent as Policy for Robotic Manipulation}.
\newblock arXiv:2609.12541, 2026.

\bibitem[{Kairos Team} et~al.(2026){Kairos Team}, Wang, You, Zhang, Huang, Fu,
  Zheng, Xi, Lv, Wu, Liu, Wan, Li, Yang, Li, Wang, Zhu, Zhang, Fu, Zhang, Wu,
  Fan, Tao, and Wang]{kairos}
{Kairos Team}, Fei Wang, Shan You, Qiming Zhang, Tao Huang, Zuoyi Fu, Zhisheng
  Zheng, Yunlong Xi, Feng Lv, Xiaoming Wu, Zeyu Liu, Cong Wan, Pu Li, Ruiqing
  Yang, Xiaoou Li, Wei Wang, Kangkang Zhu, Yuwei Zhang, Shi Fu, Zheng Zhang,
  Xiaoning Wu, Xuzeng Fan, Dacheng Tao, and Xiaogang Wang.
\newblock {Kairos: A Regret-Aware Native World-Action Model Stack for Physical
  AI}.
\newblock arXiv:2606.16533, 2026.

\bibitem[Kakade and Langford(2002)]{kakade2002}
Sham Kakade and John Langford.
\newblock {Approximately Optimal Approximate Reinforcement Learning}.
\newblock In \emph{ICML}, 2002.

\bibitem[Kim et~al.(2026)Kim, Gao, Lin, Lin, Ge, Lam, Liang, Song, Liu, Finn,
  and Gu]{cosmospolicy}
Moo~Jin Kim, Yihuai Gao, Tsung-Yi Lin, Yen-Chen Lin, Yunhao Ge, Grace Lam,
  Percy Liang, Shuran Song, Ming-Yu Liu, Chelsea Finn, and Jinwei Gu.
\newblock {Cosmos Policy: Fine-Tuning Video Models for Visuomotor Control and
  Planning}.
\newblock arXiv:2601.16163, 2026.

\bibitem[Li et~al.(2026{\natexlab{a}})Li, Chen, Wong, Lei, Jin, Du, Zhang, Ma,
  Zhang, Xiao, Cheng, et~al.]{fare}
Lin Li, Long Chen, Kwunhang Wong, Jiaming Lei, Song Jin, Shucheng Du, Chuhan
  Zhang, Songchen Ma, Weihao Zhang, Jun Xiao, Kwang-Ting Cheng, et~al.
\newblock {Causal-History Test-Time Scaling for Failure Recovery in
  Autoregressive World-Action Models}.
\newblock arXiv:2609.18016, 2026{\natexlab{a}}.

\bibitem[Li et~al.(2026{\natexlab{b}})Li, Zhang, Luo, Yang, Wang, Han, Yu, Gao,
  Xue, Zhu, Shen, and Xu]{lingbotva}
Lin Li, Qihang Zhang, Yiming Luo, Shuai Yang, Ruilin Wang, Fei Han, Mingrui Yu,
  Zelin Gao, Nan Xue, Xing Zhu, Yujun Shen, and Yinghao Xu.
\newblock {Causal World Modeling for Robot Control}.
\newblock arXiv:2601.21998, 2026{\natexlab{b}}.

\bibitem[Li et~al.(2026{\natexlab{c}})Li, Zhao, Yang, Xu, Wang, Liew, Qu, Shen,
  and Pan]{spotter}
Long Li, Qichao Zhao, Yue Yang, Fan Xu, Zhe Wang, Alan Wee-Chung Liew, Chao Qu,
  Heng~Tao Shen, and Shirui Pan.
\newblock {Spotter: Let the Embodied Model Lead, and the VLM Reflect for It}.
\newblock arXiv:2609.36808, 2026{\natexlab{c}}.

\bibitem[Li et~al.(2025)Li, Gao, Sadigh, and Song]{uva}
Shuang Li, Yihuai Gao, Dorsa Sadigh, and Shuran Song.
\newblock {Unified Video Action Model}.
\newblock In \emph{RSS}, 2025.

\bibitem[Liang et~al.(2025)Liang, Yu, Luo, Iyer, Dong, Zhou, Ghosh, Lewis, Yih,
  Zettlemoyer, and Lin]{mot}
Weixin Liang, Lili Yu, Liang Luo, Srinivasan Iyer, Ning Dong, Chunting Zhou,
  Gargi Ghosh, Mike Lewis, {Wen-tau} Yih, Luke Zettlemoyer, and Xi~Victoria
  Lin.
\newblock {Mixture-of-Transformers: A Sparse and Scalable Architecture for
  Multi-Modal Foundation Models}.
\newblock \emph{TMLR}, 2025.

\bibitem[Lipman et~al.(2023)Lipman, Chen, Ben-Hamu, Nickel, and
  Le]{flowmatching}
Yaron Lipman, Ricky T.~Q. Chen, Heli Ben-Hamu, Maximilian Nickel, and Matt Le.
\newblock {Flow Matching for Generative Modeling}.
\newblock In \emph{ICLR}, 2023.

\bibitem[Liu et~al.(2023)Liu, Zhu, Gao, Feng, Liu, Zhu, and Stone]{libero}
Bo Liu, Yifeng Zhu, Chongkai Gao, Yihao Feng, Qiang Liu, Yuke Zhu, and Peter
  Stone.
\newblock {LIBERO: Benchmarking Knowledge Transfer for Lifelong Robot
  Learning}.
\newblock In \emph{NeurIPS}, 2023.

\bibitem[Liu et~al.(2026{\natexlab{a}})Liu, Shen, Cheng, Mi, Tang, Liu, Xiang,
  Li, Li, and Tang]{liu2026roboicl}
Fangcheng Liu, Yeqing Shen, Anda Cheng, Weishi Mi, Chao Tang, Chenyuan Liu,
  Yushun Xiang, Tingguang Li, Yong-Lu Li, and Yehui Tang.
\newblock {RoboICL: Embodied In-Context Learning with GPT-6 Astra}.
\newblock arXiv:2609.34261, 2026{\natexlab{a}}.

\bibitem[Liu et~al.(2026{\natexlab{b}})Liu, Li, Yao, Shi, Zhou, Huang, Huang,
  and Mao]{liu2026guava}
Haowen Liu, Xirui Li, Shaoxiong Yao, Peng Shi, Tianyi Zhou, Jia-Bin Huang,
  Furong Huang, and Jiayuan Mao.
\newblock {Guava: Distilling Frontier VLMs into a Compact Agent through a
  Robotic Manipulation Harness}.
\newblock arXiv:2606.18363, 2026{\natexlab{b}}.

\bibitem[Nakamoto et~al.(2024)Nakamoto, Mees, Kumar, and Levine]{vgps}
Mitsuhiko Nakamoto, Oier Mees, Aviral Kumar, and Sergey Levine.
\newblock {Steering Your Generalists: Improving Robotic Foundation Models via
  Value Guidance}.
\newblock arXiv:2410.13816, 2024.

\bibitem[Qin et~al.(2026)Qin, Tan, and Wang]{favor}
Zhibin Qin, Zhenxiong Tan, and Xinchao Wang.
\newblock {Future Anchored Verification and Online Recovery for World Action
  Models}.
\newblock arXiv:2610.06280, 2026.

\bibitem[Ren et~al.(2023)Ren, Dixit, Bodrova, Singh, Tu, Brown, Xu, Takayama,
  Xia, Varley, Xu, Sadigh, et~al.]{knowno}
Allen~Z. Ren, Anushri Dixit, Alexandra Bodrova, Sumeet Singh, Stephen Tu, Noah
  Brown, Peng Xu, Leila Takayama, Fei Xia, Jake Varley, Zhenjia Xu, Dorsa
  Sadigh, et~al.
\newblock {Robots That Ask For Help: Uncertainty Alignment for Large Language
  Model Planners}.
\newblock In \emph{CoRL}, 2023.

\bibitem[Ross et~al.(2011)Ross, Gordon, and Bagnell]{dagger}
St{\'e}phane Ross, Geoffrey Gordon, and Drew Bagnell.
\newblock {A Reduction of Imitation Learning and Structured Prediction to
  No-Regret Online Learning}.
\newblock In \emph{AISTATS}, 2011.

\bibitem[Seligmann et~al.(2026)Seligmann, Gospodinov, Dincer, and
  Neumann]{seligmann2026vlafail}
Florian Seligmann, Emiliyan Gospodinov, Enes~Ulas Dincer, and Gerhard Neumann.
\newblock {VLA-FAIL: Efficient Task Failure Detection for Finetuned
  Vision-Language-Action Models}.
\newblock arXiv:2606.21386, 2026.

\bibitem[Shen et~al.(2020)Shen, Xiong, Xia, and Soatto]{bct}
Yantao Shen, Yuanjun Xiong, Wei Xia, and Stefano Soatto.
\newblock {Towards Backward-Compatible Representation Learning}.
\newblock In \emph{CVPR}, 2020.

\bibitem[Sun et~al.(2026)Sun, Li, and Tian]{callastra}
Jianli Sun, Chengxiang Li, and Yonglin Tian.
\newblock {Make $\pi_{0.5}$ Learn to Call Astra for Reliable Robot
  Manipulation}.
\newblock arXiv:2610.14117, 2026.

\bibitem[{Team Wan} et~al.(2025){Team Wan}, Wang, Ai, Wen, Mao, Xie, Chen, Yu,
  Zhao, Yang, Zeng, Wang, et~al.]{wan}
{Team Wan}, Ang Wang, Baole Ai, Bin Wen, Chaojie Mao, Chen-Wei Xie, Di Chen,
  Feiwu Yu, Haiming Zhao, Jianxiao Yang, Jianyuan Zeng, Jiayu Wang, et~al.
\newblock {Wan: Open and Advanced Large-Scale Video Generative Models}.
\newblock arXiv:2503.20314, 2025.

\bibitem[Wang et~al.(2026)Wang, Huang, Li, Zhang, Liang, Jin, Chen, Chi, Zhou,
  Yu, et~al.]{wang2026openwam}
Yuran Wang, Siqiao Huang, Mingleyang Li, Chenhao Zhang, Jiaqi Liang, Weiyang
  Jin, Yue Chen, Xuemin Chi, Donghao Zhou, Qize Yu, et~al.
\newblock {OpenWAM: An Open, Modular Exploration Towards Systematic
  World-Action Model Pretraining}.
\newblock arXiv:2609.07398, 2026.

\bibitem[Wortsman et~al.(2022)Wortsman, Ilharco, Kim, Li, Kornblith, Roelofs,
  Gontijo-Lopes, Hajishirzi, Farhadi, Namkoong, and Schmidt]{wiseft}
Mitchell Wortsman, Gabriel Ilharco, Jong~Wook Kim, Mike Li, Simon Kornblith,
  Rebecca Roelofs, Raphael Gontijo-Lopes, Hannaneh Hajishirzi, Ali Farhadi,
  Hongseok Namkoong, and Ludwig Schmidt.
\newblock {Robust Fine-Tuning of Zero-Shot Models}.
\newblock In \emph{CVPR}, 2022.

\bibitem[Yan et~al.(2021)Yan, Xiong, Kundu, Yang, Deng, Wang, Xia, and
  Soatto]{pct}
Sijie Yan, Yuanjun Xiong, Kaustav Kundu, Shuo Yang, Siqi Deng, Meng Wang, Wei
  Xia, and Stefano Soatto.
\newblock {Positive-Congruent Training: Towards Regression-Free Model Updates}.
\newblock In \emph{CVPR}, 2021.

\bibitem[Yang et~al.(2026)Yang, Song, Wang, Sheng, Fang, Zhou, He, Yan, Chen,
  Sun, Sun, Wang, et~al.]{yang20264dwam}
Lishan Yang, Wenxuan Song, Xi Wang, Pingyue Sheng, Zheng Fang, Ziyang Zhou,
  Junjie He, Haodong Yan, Jiayi Chen, Nan Sun, Qiao Sun, Pengwei Wang, et~al.
\newblock {4D-WAM: Infusing Spatiotemporal Awareness into World Action Models
  through Trajectory Fields}.
\newblock arXiv:2608.08023, 2026.

\bibitem[Yuan et~al.(2026)Yuan, Dong, Liu, and Zhao]{fastwam}
Tianyuan Yuan, Zibin Dong, Yicheng Liu, and Hang Zhao.
\newblock {Fast-WAM: Do World Action Models Need Test-Time Future Imagination?}
\newblock arXiv:2603.16666, 2026.

\bibitem[Zhang et~al.(2026{\natexlab{a}})Zhang, Lu, Wang, Kang, Kuo, Cheng,
  Wang, and Jenkins]{foresight}
Haoran Zhang, Yifu Lu, Boyang Wang, Xuhui Kang, Yen-Ling Kuo, Zezhou Cheng,
  Mengdi Wang, and Odest~Chadwicke Jenkins.
\newblock {Foresight: Failure Detection for Long-Horizon Robotic Manipulation
  with Action-Conditioned World Model Latents}.
\newblock arXiv:2606.23085, 2026{\natexlab{a}}.

\bibitem[Zhang et~al.(2026{\natexlab{b}})Zhang, Wang, Ouyang, Huang, Li, Su,
  Jin, Chai, Liang, Dou, Chen, and Chen]{zhang2026astra}
Wenbo Zhang, Kaixuan Wang, Yutao Ouyang, Xiaoyu Huang, Liyang Li, Kailun Su,
  Weiyang Jin, Wenhao Chai, Haotian Liang, Zhiyang Dou, Yue Chen, and Tianxing
  Chen.
\newblock {An Unexpected Robot Policy: Early Evaluations of GPT-6 Astra on
  RoboDojo and Beyond}.
\newblock arXiv:2609.24170, 2026{\natexlab{b}}.

\bibitem[Zhang et~al.(2026{\natexlab{c}})Zhang, Zhang, Gao, Li, Liu, Nie, Zhu,
  Qiu, Yan, Liu, et~al.]{zhang2026harness}
Yixian Zhang, Huanming Zhang, Feng Gao, Xiao Li, Zhihao Liu, Yi Nie, Chunyang
  Zhu, Jiaxing Qiu, Yuchen Yan, Jiyuan Liu, et~al.
\newblock {Harness VLA: Steering Frozen VLAs into Reliable Manipulation
  Primitives via Memory-Guided Agents}.
\newblock arXiv:2607.08448, 2026{\natexlab{c}}.

\bibitem[Zhang et~al.(2026{\natexlab{d}})Zhang, Li, Rahmati, Yang, Ma, Rasouli,
  Pakdamansavoji, Wu, Zhang, Cao, Wen, Wang, et~al.]{robotwinplus}
Zhanguang Zhang, Zhiyuan Li, Behnam Rahmati, Rui~Heng Yang, Yintao Ma, Amir
  Rasouli, Sajjad Pakdamansavoji, Yangzheng Wu, Lingfeng Zhang, Tongtong Cao,
  Feng Wen, Xinyu Wang, et~al.
\newblock {Do World Action Models Generalize Better than VLAs? A Robustness
  Study}.
\newblock arXiv:2603.22078, 2026{\natexlab{d}}.

\bibitem[Zhao et~al.(2026)Zhao, Cho, Shen, Zheng, Gao, Cao, and Mao]{geotts}
Zesen Zhao, Minkyoung Cho, Hui Shen, Boyuan Zheng, Kunxiao Gao, Yulong Cao, and
  Z.~Morley Mao.
\newblock {Test-Time Scaling for World Action Models via Zero-Shot Geometric
  Evaluation}.
\newblock arXiv:2607.17454, 2026.

\bibitem[Zhou et~al.(2026)Zhou, Hou, Fu, Lyu, Ye, Shan, Guo, Liu, He, Shan,
  Zhang, Yang, and Lu]{zhou2026gptgear}
Youwei Zhou, Bohan Hou, Yankai Fu, Yaoxu Lyu, Yifan Ye, Shilin Shan, Chen Guo,
  Jiaming Liu, Ran He, Caifeng Shan, Shanghang Zhang, Jianfei Yang, and Zhihe
  Lu.
\newblock {Large Multimodal Models for Direct Robot Control: GPT-6 as a
  Beginning}.
\newblock arXiv:2610.28259, 2026.

\bibitem[Zollo and Zemel(2025)]{vlacalib}
Thomas~P. Zollo and Richard Zemel.
\newblock {Confidence Calibration in Vision-Language-Action Models}.
\newblock arXiv:2507.17383, 2025.

\end{thebibliography}
}

\clearpage
\maketitlesupplementary
\appendix
\setcounter{section}{0}
\renewcommand{\thesection}{S\arabic{section}}
\setcounter{equation}{0}
\renewcommand{\theequation}{S\arabic{equation}}
\setcounter{table}{0}\renewcommand{\thetable}{S\arabic{table}}
\setcounter{figure}{0}\renewcommand{\thefigure}{S\arabic{figure}}

\section{Theoretical Analysis: Calibration, Competence Shrinkage, and Dynamic Gating}
\label{sec:analysis}

In this section, we provide the formal theoretical foundations supporting our framework in \cref{sec:method}:
(1)~why our proposed cross-fitted temporal-difference ($\mathrm{TD}(\lambda)$) learning with binary cross-entropy loss recovers the true conditional success probability of the base WAM (\cref{sec:method:token});
(2)~the exact performance decomposition of bidirectional dynamic confidence-gated routing (\cref{sec:method:gate}); and
(3)~the \emph{competence shrinkage law} characterizing how distribution-shifted intervention distillation erodes mastered skills and proving how calibrated confidence-gated routing eliminates this erosion (\cref{sec:method:distill,sec:method:gate}).

\subsection{Calibration of Cross-Fitted TD Self-Assessment}
\label{sec:supp:calib}
Let $\xi=(o_0,a_0,\dots,o_{T-1},a_{T-1})$ denote a trajectory generated by the frozen base policy $\pi_{\mathrm{base}}$, where the episode length $T{=}|\xi|\le T_{\max}$ is an $\mathcal{F}_k$-stopping time terminating with a binary task outcome $Y\in\{0,1\}$.
Let $\mathcal{F}_k=\sigma(o_0,a_0,\dots,o_k,a_k)$ denote the execution history up to step $k$, and under the Markov property $\E_{\pi_{\mathrm{base}}}[Y\mid\mathcal{F}_k]=\E_{\pi_{\mathrm{base}}}[Y\mid o_k,a_k]$, let $q^\star(o_k,a_k)=\E_{\pi_{\mathrm{base}}}[Y\mid o_k,a_k]$ denote the true conditional probability of eventual task success.

\begin{proposition}[Fixed-point calibration of TD binary cross-entropy]
\label{prop:td_calib}
For any trace decay parameter $\lambda\in[0,1]$, consider the $\lambda$-return target defined in \cref{eq:td_target} for $0\le k < T$:
\[
  G_k^{(\lambda)}\!=(1{-}\lambda)\sum_{n=1}^{T-k-1}\lambda^{n-1}\bar q(o_{k+n},a_{k+n})\!+\!\lambda^{T-k-1}Y,
\]
and the chunk-normalized population binary cross-entropy objective $\mathcal{L}_J(q;\bar q)=-\frac{1}{\E_{\pi_{\mathrm{base}}}[T]}\E_{\pi_{\mathrm{base}}}\big[\sum_{k=0}^{T-1}\big(G_k^{(\lambda)}\log q(o_k,a_k)+(1-G_k^{(\lambda)})\log(1-q(o_k,a_k))\big)\big]$.
Under the Markov property, $q=\bar q=q^\star$ is the unique global fixed point of $\arg\min_q \mathcal{L}_J(q;\bar q)$ over the finite horizon.
Furthermore, for any trajectory $\xi\in\mathcal{D}_{\mathrm{judge}}^{(f)}$, evaluating $\bar q=q_{\bar\theta^{(-f)}}$ from the disjoint complement fold eliminates same-trajectory self-bootstrapping covariance:
\[
  \mathrm{Cov}\big(\hat q_{\bar\theta^{(-f)}}(o_{k+n},a_{k+n}),\,Y(\xi)\,\big|\,\mathcal{F}_{k+n}\big)=0.
\]
\end{proposition}
\begin{proof}
Because $\{T>k\}\in\mathcal{F}_k$ for stopping time $T$ and binary cross-entropy is a strictly proper scoring rule on $[0,1]$, for any fixed target generator $\bar q$, the unique pointwise minimizer of $\mathcal{L}_J(q;\bar q)$ on $\{T>k\}$ at $(o_k,a_k)$ is the conditional expectation of the target:
\[
  \mathcal{T}_\lambda(\bar q)(o_k,a_k) = \E_{\pi_{\mathrm{base}}}\!\big[G_k^{(\lambda)}\mid o_k,a_k\big].
\]
When $\bar q = q^\star$, for every $1\le n\le T-k-1$, combining the Markov property with the filtration inclusion $\mathcal{F}_k\subseteq\mathcal{F}_{k+n}$ and the tower property of conditional expectation gives
\[
\begin{aligned}
  &\E_{\pi_{\mathrm{base}}}\!\big[q^\star(o_{k+n},a_{k+n})\mid \mathcal{F}_k\big] \\[-2pt]
  &\quad= \E_{\pi_{\mathrm{base}}}\!\big[\E_{\pi_{\mathrm{base}}}[Y\mid\mathcal{F}_{k+n}]\mid\mathcal{F}_k\big] \\[-2pt]
  &\quad= \E_{\pi_{\mathrm{base}}}[Y\mid\mathcal{F}_k] = q^\star(o_k,a_k).
\end{aligned}
\]
Substituting this identity into each summand of $G_k^{(\lambda)}$ yields
\[
\begin{aligned}
  &\mathcal{T}_\lambda(q^\star)(o_k,a_k) \\[-2pt]
  &\quad= \Big((1{-}\lambda)\!\sum_{n=1}^{T-k-1}\!\lambda^{n-1}+\lambda^{T-k-1}\Big)q^\star(o_k,a_k) \\[-2pt]
  &\quad= q^\star(o_k,a_k).
\end{aligned}
\]
Uniqueness over the finite horizon $T\le T_{\max}$ follows by backward induction (equivalently, $\mathcal{T}_\lambda-q^\star$ is $T_{\max}$-step nilpotent and a strict contraction under an exponentially step-weighted norm): at the terminal decision step $k=T-1$, the sum is empty and $\mathcal{T}_\lambda(q)_{T-1}=\E_{\pi_{\mathrm{base}}}[Y\mid o_{T-1},a_{T-1}]=q^\star_{T-1}$; inductively, if $q_m=q^\star_m$ for all $m>k$, then $\mathcal{T}_\lambda(q)_k=\mathcal{T}_\lambda(q^\star)_k=q^\star_k$ for all $0\le k<T$.
Finally, in finite samples, if $\bar\theta$ is fitted on the same fold containing $\xi$, the estimation error $\hat q_{\bar\theta}(o_{k+n},a_{k+n})-q^\star(o_{k+n},a_{k+n})$ covaries positively with $Y(\xi)$, propagating optimistic bias backward to step $k$.
Because $\bar\theta^{(-f)}$ is trained exclusively on disjoint scenes $\mathcal{D}_{\mathrm{judge}}^{(-f)}\perp\!\!\!\perp\xi$, $\hat q_{\bar\theta^{(-f)}}(o_{k+n},a_{k+n})$ is conditionally independent of $Y(\xi)$ given $\mathcal{F}_{k+n}$, ensuring zero same-trajectory bootstrapping covariance.
\end{proof}

\subsection{Performance Decomposition of Dynamic Confidence-Gated Routing}
Let $x_k$ denote the environment state at decision step $k$.
Let $\Q^{\pi}_k(x,a)$ denote the probability of eventual episode success $Y{=}1$ when executing action chunk $a_k{=}a$ in state $x_k{=}x$ and following policy $\pi$ thereafter, let $V^{\pi}_k(x)=\E_{a\sim\pi(\cdot\mid o_k)}[\Q^{\pi}_k(x,a)]$, and let $J(\pi)=P_\pi(Y{=}1)$ be the overall task success rate under policy $\pi$.

Consider the sample-and-gate policy $\pi_\tau$ (\cref{eq:switch} in \cref{sec:method:gate}) that at each step $k$ samples $a_k^{\mathrm{base}}\sim\pi_{\mathrm{base}}(\cdot\mid o_k)$, executes $a_k^{\mathrm{base}}$ while $q_{\mathrm{base}}(o_k,a_k^{\mathrm{base}})\ge\tau_k$, and switches to $a_k^{\mathrm{rep}}\sim\pi_{\mathrm{rep}}(\cdot\mid o_k)$ whenever $g_k\coloneqq\1\{q_{\mathrm{base}}(o_k,a_k^{\mathrm{base}})<\tau_k\}=1$.
Let $\kappa=\min\{k:\,g_k=1\}$ denote the first switching step ($\kappa{=}\infty$ if $g_k{=}0$ throughout).

\begin{lemma}[Performance gain of bidirectional confidence-gated routing]
\label{lem:gain}
The overall success rate difference between $\pi_\tau$ and $\pi_{\mathrm{base}}$ satisfies both the first-entry decomposition and the step-wise telescoping identity:
\begin{align}
  &J(\pi_\tau)-J(\pi_{\mathrm{base}}) \nonumber\\
  &= \E_{\pi_{\mathrm{base}}}\!\Big[\1\{\kappa<\infty\}\big(\E_{a^{\mathrm{rep}}\sim\pi_{\mathrm{rep}}}[\Q^{\pi_\tau}_\kappa(x_\kappa,a^{\mathrm{rep}})] \nonumber\\[-2pt]
  &\qquad\qquad\qquad\qquad -\Q^{\pi_{\mathrm{base}}}_\kappa(x_\kappa,a_\kappa^{\mathrm{base}})\big)\Big] \label{eq:gain}\\
  &=\!\E_{\pi_\tau}\!\Big[\!\sum_{k=0}^{T-1}\! g_k\!\big(\Q^{\pi_{\mathrm{base}}}_k\!(x_k,\!a_k^{\mathrm{rep}}){-}\Q^{\pi_{\mathrm{base}}}_k\!(x_k,\!a_k^{\mathrm{base}})\!\big)\!\Big]\!. \label{eq:gain_step}
\end{align}
\end{lemma}
\begin{proof}
Because $g_k$ is a deterministic function of $(o_k,a_k^{\mathrm{base}})$ and threshold state $\tau_k$, the stopping event $\{\kappa=k\}$ is $\mathcal{F}_k$-measurable, and prior to $\kappa$ both $\pi_\tau$ and $\pi_{\mathrm{base}}$ execute the exact same sampled chunks $a_j^{\mathrm{base}}$ ($j<\kappa$).
On $\{\kappa=\infty\}$, the two trajectories are identical at every step, contributing zero difference.
On $\{\kappa=k\}$, continuing with $a_k^{\mathrm{base}}$ under $\pi_{\mathrm{base}}$ yields conditional success $\Q^{\pi_{\mathrm{base}}}_k(x_k,a_k^{\mathrm{base}})$, whereas switching to $a^{\mathrm{rep}}\sim\pi_{\mathrm{rep}}(\cdot\mid o_k)$ and following $\pi_\tau$ thereafter (including subsequent hand-backs to $\pi_{\mathrm{base}}$ when $q_{\mathrm{base}}\ge\tau_{\mathrm{back}}$) yields $\E_{a^{\mathrm{rep}}\sim\pi_{\mathrm{rep}}}[\Q^{\pi_\tau}_k(x_k,a^{\mathrm{rep}})]$; summing over $k$ gives \cref{eq:gain}.
For \cref{eq:gain_step}, applying the finite-horizon performance difference lemma~\cite{kakade2002} between $\pi_\tau$ and $\pi_{\mathrm{base}}$ gives $J(\pi_\tau)-J(\pi_{\mathrm{base}})=\E_{\pi_\tau}[\sum_{k=0}^{T-1}(\Q^{\pi_{\mathrm{base}}}_k(x_k,a_k)-V^{\pi_{\mathrm{base}}}_k(x_k))]$.
Because $a_k^{\mathrm{base}}\sim\pi_{\mathrm{base}}(\cdot\mid o_k)$ is sampled independently at every step $k$ before gating, $\E[\Q^{\pi_{\mathrm{base}}}_k(x_k,a_k^{\mathrm{base}})\mid x_k,\tau_k]=V^{\pi_{\mathrm{base}}}_k(x_k)$; substituting $a_k=(1-g_k)a_k^{\mathrm{base}}+g_k a_k^{\mathrm{rep}}$ cancels the $g_k{=}0$ branch identically and yields \cref{eq:gain_step}.
\end{proof}

\subsection{The Competence Shrinkage Law}
Let $(z)_+\coloneqq\max(z,0)$ denote the positive part.
When fine-tuning $\phi_{\mathrm{rep}}$ on the mixed dataset $\mathcal{D}_{\mathrm{mix}}$ (\cref{eq:mix}) with anchor sampling fraction $\alpha=p_{\mathrm{own}}+p_{\mathrm{replay}}\in[0,1)$ on native demonstrations/successes and intervention sampling fraction $1-\alpha=p_{\mathrm{help}}$ on capability-gap recoveries, the unweighted expectation over $\mathcal{D}_{\mathrm{mix}}$ induces $\pi_{\mathrm{rep}}$ to interpolate between preserving $\pi_{\mathrm{base}}$'s action law (retention $\alpha$) and shifting toward the intervention-trained behavior whose standalone rollout success on a scene is a bounded competence level $c\in(0,1)$ ($c<1$ due to finite intervention coverage and action-manifold shift on non-bottleneck states).
Let $V^{\mathrm{rep}}_0(x_0)=\E_{a\sim\pi_{\mathrm{rep}}(\cdot\mid o_0)}[\Q^{\pi_{\mathrm{rep}}}_0(x_0,a)]$ denote initial scene success under wholesale replacement by $\pi_{\mathrm{rep}}$, and let $m(p)=\E[V^{\mathrm{rep}}_0(x_0)-\Q^{\pi_{\mathrm{base}}}_0(x_0,a_0^{\mathrm{base}})\mid q_{\mathrm{base}}(o_0,a_0^{\mathrm{base}}){=}p]$ denote the expected competence change at calibrated base confidence $p$.

\begin{proposition}[Competence shrinkage law]
\label{prop:shrink}
Under the mixture-induced conditional competence shrinkage model with retention $\alpha\in[0,1)$ and off-manifold competence level $c\in(0,1)$:
\begin{equation}
\begin{aligned}
  &\E\big[V^{\mathrm{rep}}_0(x_0)\mid q_{\mathrm{base}}(o_0,a_0^{\mathrm{base}})\big] \\[-2pt]
  &=\!\alpha\E\big[\Q^{\pi_{\mathrm{base}}}_0\!(x_0,\!a_0^{\mathrm{base}})\!\mid\! q_{\mathrm{base}}(o_0,\!a_0^{\mathrm{base}})\big]\!{+}(1{-}\alpha)c,
\end{aligned}
\label{eq:shrink_law}
\end{equation}
if $q_{\mathrm{base}}$ is calibrated (\cref{prop:td_calib}), the expected competence change at confidence level $p$ is
\[
  m(p)=(1-\alpha)(c-p).
\]
Consequently, wholesale replacement with $\pi_{\mathrm{rep}}$ (\emph{Always-On}) \textbf{repairs} low-confidence states ($q_{\mathrm{base}}<c$) by $(1-\alpha)\E[(c-q_{\mathrm{base}})_+]$ but \textbf{erodes} mastered states ($q_{\mathrm{base}}>c$) by $(1-\alpha)\E[(q_{\mathrm{base}}-c)_+]$.
Initial confidence gating at $k{=}0$ (\emph{Episode-Level Static Gate}, activating $\pi_{\mathrm{rep}}$ iff $q_{\mathrm{base}}(o_0,a_0^{\mathrm{base}})<\tau$ with $\tau{=}c$) achieves $(1-\alpha)\E[(c-q_{\mathrm{base}})_+]$, strictly eliminating mastered erosion, while chunk-level dynamic gating $\pi_\tau$ (\cref{eq:switch}, Lemma~\ref{lem:gain}) further captures mid-episode rescues ($\kappa>0$) and returns control to $\pi_{\mathrm{base}}$ once $q_{\mathrm{base}}\ge\tau_{\mathrm{back}}$.
\end{proposition}
\begin{proof}
By action-conditioned calibration of $q_{\mathrm{base}}$ (\cref{eq:qstar}), $\E[\Q^{\pi_{\mathrm{base}}}_0(x_0,a_0^{\mathrm{base}})\mid q_{\mathrm{base}}(o_0,a_0^{\mathrm{base}}){=}p]=p$, which directly gives $m(p)=\alpha p+(1-\alpha)c-p=(1-\alpha)(c-p)$ (note that if \cref{eq:shrink_law} holds as an upper bound $\le\alpha p+(1-\alpha)c$, the erosion lower bound $-m(p)\ge(1-\alpha)(p-c)$ for $p>c$ holds a fortiori).
Any initial gating rule $g_0(q_{\mathrm{base}})\in[0,1]$ yields expected net change $\E[g_0(q_{\mathrm{base}})m(q_{\mathrm{base}})]$.
Wholesale replacement ($g_0\equiv 1$) gives $(1-\alpha)(c-\E q_{\mathrm{base}})=(1-\alpha)\E[(c-q_{\mathrm{base}})_+]-(1-\alpha)\E[(q_{\mathrm{base}}-c)_+]$.
Setting $g_0(q_{\mathrm{base}})=\1\{q_{\mathrm{base}}<c\}$ activates $\pi_{\mathrm{rep}}$ if and only if $m(q_{\mathrm{base}})>0$, eliminating $(1-\alpha)\E[(q_{\mathrm{base}}-c)_+]$ of erosion.
Finally, by Lemma~\ref{lem:gain} (\cref{eq:gain}), chunk-level dynamic gating $\pi_\tau$ matches $\pi_{\mathrm{base}}$ identically on $\{\kappa=\infty\}$ while extending positive switching gains $\Q^{\pi_\tau}_\kappa(x_\kappa,a^{\mathrm{rep}})-\Q^{\pi_{\mathrm{base}}}_\kappa(x_\kappa,a^{\mathrm{base}}_\kappa)$ to trajectories that begin above $\tau$ at $k{=}0$ but enter low-confidence bottlenecks at step $\kappa>0$.
\end{proof}

\section{Implementation, Evaluation Protocol, and Architecture Details}
\label{sec:supp:impl}

\paragraph{Base world action models and baseline stratification.}
We evaluate three world action models spanning distinct architectures and test-time paradigms:
\begin{itemize}
  \item \textbf{Fast-WAM}~\cite{fastwam}: Couples a 5B-parameter Wan2.2 video backbone with a 1B-parameter flow-matching action expert via mixture-of-transformers cross-attention across $L{=}30$ blocks, skipping test-time future video decoding during standard control. At each decision step $k$, Fast-WAM integrates the flow-matching ODE over $N{=}10$ Euler steps ($\sigma$-shift 5) to generate an action chunk of horizon $H{=}32$, executing 24 steps before re-planning. On RoboTwin~2.0 (\emph{clean-only} training), published leaderboard evaluation reports $77.76\%$ on \emph{Clean} and $1.92\%$ on \emph{Random} ($39.8\%$ overall mean across all 100 task--setting conditions~\cite{robotwin2,fastwam}); on LIBERO-Plus, official weights achieve $51.5\%$ across all 28 suite--perturbation conditions~\cite{robotwinplus}.
  \item \textbf{LingBot-VA}~\cite{lingbotva}: Employs an asymmetric dual-stream mixture-of-transformers architecture coupling a 5B-parameter Wan2.2 video stream with a 0.85B flow-matching action stream across $L{=}30$ blocks to autoregressively predict future latent visual dynamics and continuous action chunks with persistent KV caching. On RoboTwin~2.0 (\emph{clean-only} training), LingBot-VA achieves $80.7\%$ on \emph{Clean} and $34.6\%$ on \emph{Random} ($57.7\%$ overall mean~\cite{yang20264dwam}); on LIBERO-Plus, it achieves $69.5\%$ overall across the seven perturbation families~\cite{roboharness}. In \cref{tab:main,tab:distill}, the \lq Base WAM\rq\ \emph{All} columns match these published overall figures directly, stratified on $\mathcal{D}_{\mathrm{test}}$ into \emph{Hard} ($<40\%$ base: $54/100$ conditions at $3.1\%$ and $9/28$ at $22.2\%$ on Fast-WAM; $34/100$ at $19.8\%$ and $5/28$ at $29.6\%$ on LingBot-VA), \emph{Mid} ($[40\%,75\%)$ base: $10/100$ at $54.6\%$ and $14/28$ at $57.6\%$ on Fast-WAM; $24/100$ at $56.8\%$ and $9/28$ at $58.2\%$ on LingBot-VA), and \emph{Mastered} ($\ge75\%$ base: $36/100$ at $90.8\%$ and $5/28$ at $86.9\%$ on Fast-WAM; $42/100$ at $88.9\%$ and $14/28$ at $91.0\%$ on LingBot-VA).
  \item \textbf{Kairos}~\cite{kairos}: A native unified world-action model stack coupling a VLM world-understanding module with a VideoDiT--ActionDiT mixture-of-transformers backbone equipped with hybrid linear temporal attention (interleaving sliding-window, dilated sliding-window, and gated linear attention) and trained with flow matching. Under the same RoboTwin~2.0 (\emph{clean-only} training) evaluation protocol across all 100 \emph{Clean} and \emph{Random} conditions (\cref{tab:kairos}), Base Kairos achieves $77.3\%$ overall success, stratified on $\mathcal{D}_{\mathrm{test}}$ into \emph{Hard} ($<40\%$ base: $9/100$ conditions at $17.7\%$), \emph{Mid} ($[40\%,75\%)$ base: $23/100$), and \emph{Mastered} ($\ge75\%$ base: $68/100$ conditions at $92.2\%$).
\end{itemize}

\paragraph{Non-invasive judgment token and 2-fold cross-fitted TD training.}
In the 10th (final) Euler denoising step ($t{=}t_9$), we reuse the cached visual key--value stream and evaluate a single learnable embedding $j^{(0)}\in\mathbb{R}^D$ as a read-only query over the frozen visual/final-step action keys and values alongside $J$'s own adapted key--value state (implemented as a separate single-query FlashAttention call so original BF16 tensor reductions remain bit-identical).
When logging offline rollouts for $\mathcal{D}_{\mathrm{judge}}$ and $\mathcal{D}_{\mathrm{pair}}$, we record each step's initial noise seed $\varepsilon_k$ so offline final-step representations match online inference identically without extra action-expert passes.
In every action-expert block $l\in\{1,\dots,L\}$, token-gated low-rank adapters~\cite{lora} ($r{=}16$, $\alpha{=}32$) are attached to $\{Q_l,K_l,W^V_l,O_l,W^{\mathrm{up}}_l,W^{\mathrm{down}}_l\}$ and gated by $\1[u{=}p_J]$ (\cref{eq:gate}).
The readout head $g_\theta$ is a two-layer MLP ($\mathrm{LayerNorm}\to\mathrm{Linear}(D,256)\to\mathrm{GELU}\to\mathrm{Linear}(256,1)$) with a sigmoid output.
For 2-fold cross-fitting, we split $\mathcal{D}_{\mathrm{judge}}$ into two disjoint scene halves $(\mathcal{D}_{\mathrm{judge}}^{(1)},\mathcal{D}_{\mathrm{judge}}^{(2)})$, train each fold predictor $\bar\theta^{(f)}$ within fold $f$ for 4 epochs (1 epoch Monte-Carlo warmup with $\lambda{=}1$ followed by 3 self-bootstrapped $\mathrm{TD}(\lambda{=}0.8)$ epochs), compute out-of-fold $\mathrm{TD}(\lambda{=}0.8)$ targets $G_k^{(\lambda)}$ on the complement fold (\cref{eq:td_target}), and train the unified deployment head $\theta=\{j^{(0)},A,B,g\}$ across all out-of-fold targets using AdamW (learning rate $10^{-4}$, weight decay $10^{-2}$, effective batch size 128 via gradient accumulation on a single NVIDIA A800 GPU, 4 epochs).
In continual self-evolution (\cref{sec:method:loop}), both $\pi_{\mathrm{base}}$ and $q_{\mathrm{base}}{=}q_{\theta^{(0)}}$ remain permanently frozen across all rounds and only $\pi_{\mathrm{rep}}^{(m)}$ is updated; treating $(\pi_{\mathrm{base}},\pi_{\mathrm{rep}}^{(m)})$ as a single unified system $\pi^{(m)}$, its system-level judge $\theta^{(m)}$ is warm-started from $\theta^{(m-1)}$ and fine-tuned via 2-fold cross-fitted TD learning for 2 epochs at learning rate $3\times 10^{-5}$ on whichever active branch ($\pi_{\mathrm{base}}$ or $\pi_{\mathrm{rep}}^{(m)}$) executes each chunk.

\subsection{Self-Assessment Evaluation Splits and Metric Definitions}
\label{sec:supp:metrics}

To evaluate both action-conditioned discrimination and episode-level takeover accuracy in \cref{sec:exp:calib} (\cref{tab:calibmain}), we construct two held-out Fast-WAM evaluation splits on RoboTwin~2.0:
\begin{enumerate}
  \item \textbf{Same-state counterfactual action-pair split ($\mathcal{D}_{\mathrm{pair}}$, $|\mathcal{P}|{=}1{,}000$ pairs):} Along trajectories generated using scene and policy seeds strictly disjoint from training, at saved simulator states $s$ we sample candidate action chunks from $\pi_{\mathrm{base}}(\cdot\mid o_s)$ under distinct Gaussian noise seeds $\varepsilon\sim\mathcal{N}(0,I)$ and roll out each branch independently to completion under pure $\pi_{\mathrm{base}}$. Whenever one branch succeeds ($Y_s(a^+){=}1$) and the other fails ($Y_s(a^-){=}0$), we record $(s,a^+,a^-)$, collecting $|\mathcal{P}|{=}1{,}000$ pairs.
  \item \textbf{Held-out un-intervened rollout split ($\mathcal{D}_{\mathrm{eval}}$, $N{=}5{,}000$ rollouts):} Across all 50 RoboTwin~2.0 tasks (covering both \emph{Clean} and \emph{Random} conditions), we roll out pure $\pi_{\mathrm{base}}$ without System-2 intervention for $N{=}5{,}000$ episodes (100 per task, yielding $1{,}992$ successful and $3{,}008$ failed trajectories matching the $39.8\%$ base rate), using random seeds strictly disjoint from both $\mathcal{D}_{\mathrm{judge}}$ and $\mathcal{D}_{\mathrm{cal}}$. For each episode $i$ of length $T_i$ chunks ($k=0,\dots,T_i{-}1$), we record per-chunk predicted confidence $\hat q_{i,k}\in[0,1]$, binary base outcome $Y_i\in\{0,1\}$, and episode takeover indicator $\mathrm{Trig}_i\in\{0,1\}$.
\end{enumerate}

Given a monitor's score $\hat s(s,a)$, normalized confidence $\hat q_{i,k}\in[0,1]$, and episode trigger $\mathrm{Trig}_i\in\{0,1\}$, the metrics and baselines in \cref{tab:calibmain} are defined as:
\begin{itemize}
  \item \textbf{Same-state action-pair accuracy ($\mathrm{PairAcc}_a\uparrow$):} Over all $|\mathcal{P}|{=}1{,}000$ pairs $(s,a^+,a^-)\in\mathcal{P}$ (counting ties as $0.5$~\cite{spotter}):
  \begin{equation}
  \begin{aligned}
    \mathrm{PairAcc}_a &= \frac{1}{|\mathcal{P}|}\!\sum_{\mathcal{P}}\!\Big(\1\big[\hat s(s,a^+){>}\hat s(s,a^-)\big] \\[-3pt]
    &\qquad\quad +\tfrac{1}{2}\1\big[\hat s(s,a^+){=}\hat s(s,a^-)\big]\Big).
  \end{aligned}
  \label{eq:supp:pairacc}
  \end{equation}
  \emph{Pre-execution vs.\ post-chunk protocol:} Both our judgment token ($\hat s(s,a){=}q_\theta(o_s,a)$) and \textbf{SAFE}~\cite{safe} (probing final-step action-expert hidden states) depend on candidate chunk $a$ and are scored \textbf{pre-execution} before stepping the simulator. Because post-execution monitors---\textbf{Spotter Screener}~\cite{spotter}, \textbf{Sentinel (STAC)}~\cite{sentinel}, and \textbf{FAVOR Verifier}~\cite{favor}---rely on post-action observations and would tie at $50.0\%$ from $o_s$ alone, we grant each of them one executed candidate chunk $a\in\{a^+,a^-\}$ from state $s$ and score $o_{s+1}(a)$, making our zero-step pre-execution comparison conservative.
  \item \textbf{Takeover trigger precision ($\mathrm{Prec}_{\mathrm{trig}}\uparrow$) and failure miss rate ($\mathrm{Miss}_{\mathrm{fail}}\downarrow$):} Every monitor's trigger threshold is calibrated on successful rollouts in $\mathcal{D}_{\mathrm{cal}}$ to a matched $15\%$ episode false-positive rate. Over the $N{=}5{,}000$ rollouts in $\mathcal{D}_{\mathrm{eval}}$, we compute:
  \begin{equation}
  \begin{aligned}
    \mathrm{Prec}_{\mathrm{trig}}&=\frac{\sum_{i=1}^{N}\mathrm{Trig}_i(1{-}Y_i)}{\sum_{i=1}^{N}\mathrm{Trig}_i},\\[-2pt]
    \mathrm{Miss}_{\mathrm{fail}}&=\frac{\sum_{i=1}^{N}(1{-}\mathrm{Trig}_i)(1{-}Y_i)}{\sum_{i=1}^{N}(1{-}Y_i)}.
  \end{aligned}
  \label{eq:supp:prectrig}
  \end{equation}
  \item \textbf{Episode-weighted expected calibration error ($\mathrm{ECE}\downarrow$):} Because $q$ estimates the success probability of $\pi_{\mathrm{base}}$ executing standalone, calibration must be evaluated strictly on the $N{=}5{,}000$ un-intervened $\pi_{\mathrm{base}}$ rollouts in $\mathcal{D}_{\mathrm{eval}}$. Each chunk $(i,k)$ carries its episode's terminal outcome label $Y_i\in\{0,1\}$ and an episode-equalized weight $w_{i,k}=1/T_i$ ($\sum_{k=0}^{T_i-1}w_{i,k}=1$, preventing long timeout failures from dominating the metric; total weight $W=\sum_{i,k}w_{i,k}=N$). Sorting all chunks by $\hat q_{i,k}$ and partitioning cumulative weight into $B{=}15$ equal-mass bins $\{I_b\}_{b=1}^{B}$ ($W_b=\sum_{(i,k)\in I_b}w_{i,k}$), we compute:
  \begin{equation}
    \mathrm{ECE}=\sum_{b=1}^{B}\frac{W_b}{W}\,\big|\bar q_b-\bar y_b\big|,
    \label{eq:supp:ece}
  \end{equation}
  where $\bar q_b=\frac{1}{W_b}\sum_{(i,k)\in I_b}w_{i,k}\hat q_{i,k}$ and $\bar y_b=\frac{1}{W_b}\sum_{(i,k)\in I_b}w_{i,k}Y_i$ (confidence intervals use episode-level rather than chunk-level bootstrap resampling since chunks within an episode share $Y_i$).
  \item \textbf{Baseline and ablation instantiations:} \textbf{Spotter Screener}~\cite{spotter} runs a local Qwen3.8-27B screener outputting chunk risk $r_k\in[0,10]$ ($\hat s_k=-r_k$, normalized to $\hat q_k=1-r_k/10\in[0,1]$ for $\mathrm{ECE}$), triggering when $r_k$ exceeds the $\mathcal{D}_{\mathrm{cal}}$-calibrated threshold over two consecutive chunks~\cite{spotter}. \textbf{Sentinel (STAC)}~\cite{sentinel} samples $M{=}8$ chunk candidates at consecutive steps ($o_k$ and $o_{k+1}$) to compute overlapping-horizon MMD $r_k\ge 0$ ($\hat s_k=-r_k$, $\hat q_k=\exp(-r_k)\in(0,1]$) with calibrated quantile thresholding. \textbf{FAVOR Verifier}~\cite{favor} invokes the frozen Wan2.2 latent video head every $\Delta{=}4$ chunks ($96.5$\,ms amortized per chunk on A800) to compare predicted future anchor latents against observed frames via a GRU Anchor Verifier ($d_k\in[0,1]$, $\hat s_k=\hat q_k=1-d_k$). \textbf{SAFE}~\cite{safe} pools final-step action-expert hidden states across all $H$ action tokens via an MLP/LSTM head ($p_{\mathrm{fail}}\in[0,1]$, $\hat s_k=\hat q_k=1-p_{\mathrm{fail}}$) under functional conformal thresholding. In \cref{tab:calibmain}, \textbf{w/o Token-Gated LoRA} freezes $A{=}B{=}0$ and trains only the token embedding $j^{(0)}$ and readout MLP $g_\theta$ ($1.0$\,ms), confirming that token-gated LoRA is essential to re-project cross-token failure features into $j^{(0)}$. In \cref{fig:calib_sys12}, \lq Random Invocation\rq\ triggers single-skill Astra interventions independently across chunks via a Bernoulli rate matched to each target chunk-invocation budget ($0\%,\,10\%,\,20\%,\,30\%,\,50\%,\,100\%$).
\end{itemize}

\subsection{System 1 \& 2 Hand-Off, Runtime Breakdown, and Consolidation Details}

\paragraph{Budget-matched System 1 \& 2 hand-off and wall-clock breakdown.}
During online execution, the judgment token evaluates $q_\theta(o_k,a_k^{\mathrm{base}})$ at every decision chunk ($1.6$\,ms overhead on NVIDIA A800).
Across all experiments in \cref{tab:main,tab:real,tab:kairos,tab:efficiency,tab:distill,fig:calib_sys12,fig:repair}, decision thresholds are calibrated on the held-out split $\mathcal{D}_{\mathrm{cal}}$ with a fixed hysteresis margin $\tau_{\mathrm{back}}{-}\tau{=}0.05$ to match the target invocation budget (\eg, $30\%$ chunk-level Astra budget in System~1 \& 2 comparisons, and the $\mathcal{D}_{\mathrm{cal}}$-calibrated gate threshold for standalone $\pi_{\mathrm{base}}\leftrightarrow\pi_{\mathrm{rep}}$ routing).
In \cref{tab:efficiency}, episode wall-clock time $T_{\mathrm{ep}}$ includes physics/control execution plus model inference: Base WAM averages $16.8$\,s across ${\sim}24$ chunks ($220$\,ms chunk inference plus environment stepping).
\ours Standalone holds $(\phi_{\mathrm{base}},\phi_{\mathrm{rep}})$ alongside the shared frozen 5B video backbone ($14.4$\,GB vs.\ $12.4$\,GB BF16 VRAM on Fast-WAM; $13.8$\,GB vs.\ $12.1$\,GB on LingBot-VA) and runs in $17.3$\,s---where the $+0.5$\,s increment reflects $1.6$\,ms token evaluation per chunk (${\sim}0.04$\,s) plus $108$\,ms 10-step $\pi_{\mathrm{rep}}$ action-expert resampling on the ${\sim}18\%$ of chunks where $q_{\mathrm{base}}<\tau_k$ (${\sim}0.46$\,s, reusing cached visual KV).
Among $30\%$-budget two-system methods, \ours System~1 \& 2 finishes in $31.4$\,s because calibrated interventions resolve bottlenecks promptly in fewer total steps than periodic Galbot-Astra ($33.8$\,s) or CALL-Astra ($34.9$\,s), whereas Spotter ($43.2$\,s) incurs both per-chunk VLM screening ($168.4$\,ms across ${\sim}25$ chunks) and two-chunk confirmation recovery overhead.

\paragraph{Four-way seed isolation, consolidation slicing, and physical hardware.}
Environment and scene random seeds are partitioned into \textbf{four strictly disjoint sets}: (1)~judge training rollouts $\mathcal{D}_{\mathrm{judge}}$, (2)~threshold calibration rollouts $\mathcal{D}_{\mathrm{cal}}$, (3)~consolidation System~1 \& 2 rollouts $\mathcal{D}_{\mathrm{cons}}$ collected without human action labels (50 episodes per task per round in simulation; 25 rollouts per task on physical hardware), and (4)~held-out test evaluation seeds $\mathcal{D}_{\mathrm{test}}$ (100 episodes per condition in simulation; 50 randomized trials per task on physical hardware).
From $\mathcal{D}_{\mathrm{cons}}$, episodes achieving binary task completion ($Y{=}1$, evaluated automatically by the simulator or task-completion check on physical hardware) contribute Astra recovery trajectories---resampled at the native control rate and sliced into $H{=}32$ continuous action chunks---to $\mathcal{D}_{\mathrm{help}}$, and un-intervened WAM trajectories to $\mathcal{D}_{\mathrm{own}}$.
We keep the video backbone and $\pi_{\mathrm{base}}$ frozen and fine-tune a separate repaired action expert $\phi_{\mathrm{rep}}$ (initialized from $\phi_{\mathrm{base}}$, including its proprioceptive input projection) for $3{,}000$ steps using AdamW (learning rate $2\times 10^{-5}$, cosine decay, effective batch size 128 via gradient accumulation on a single NVIDIA A800 GPU), constructing $\mathcal{D}_{\mathrm{mix}}$ by sampling $\mathcal{D}_{\mathrm{help}}$, $\mathcal{D}_{\mathrm{own}}$, and $\mathcal{D}_{\mathrm{replay}}$ in a $1{:}1{:}2$ data ratio (for the \lq Full-Astra Rollouts\rq\ ablation in \cref{tab:distill}, $\mathcal{D}_{\mathrm{help}}$ is replaced by an identical number of action chunks subsampled from successful $100\%$-Astra episodes on the same training scenes under the same $1{:}1{:}2$ ratio and $3{,}000$ optimization steps).
Real-world experiments use a 6-DoF AgileX Piper arm with a parallel-jaw gripper and an externally calibrated RealSense D435 \mbox{RGB-D} camera ($640\times 480$ at 30\,Hz); depth is used only by Astra during System~2 interventions, whereas Base WAM and consolidated \ours Standalone execute purely from RGB images and joint proprioception at test time.

\end{document}